\documentclass[journal]{vgtc}  
\usepackage{graphicx}
\graphicspath{{Figures/}}
\DeclareGraphicsExtensions{.pdf,.png,.jpg,.eps}
\usepackage{times}                  
    
\usepackage{booktabs} 
\usepackage{lipsum}       
\usepackage{mwe}                      
\usepackage{newtxmath}
\usepackage{amsmath}
\usepackage{multirow} 
\usepackage{makecell}
\usepackage{booktabs}
\usepackage{colortbl}
\usepackage{tabularx}
\usepackage{mathtools}
\usepackage{textcomp}
\usepackage{siunitx}
\usepackage{algorithmic}
\usepackage{listings}
\usepackage{multirow}
\usepackage{boldline}
\usepackage{array}
\newcolumntype{C}[1]{>{\centering\arraybackslash}p{#1}}
\newcolumntype{M}[1]{>{\centering\arraybackslash}m{#1}}
\newcolumntype{L}[1]{>{\raggedright\arraybackslash}m{#1}} 
\usepackage{caption}
\usepackage[linesnumbered,ruled,lined,commentsnumbered]{algorithm2e}
\usepackage[percent]{overpic}
\usepackage[table]{xcolor}
\definecolor{lightgreen}{RGB}{200, 255, 200}
\definecolor{darkgreen}{RGB}{120, 200, 120}
\definecolor{darkgreen1}{RGB}{50, 120, 50}
\definecolor{lightred}{RGB}{255, 200, 200}
\definecolor{lightyellow}{RGB}{255, 255, 200}
\definecolor{lightyellow1}{RGB}{255, 220, 80}

\usepackage[labelformat=simple]{subcaption}

\usepackage{url}
\usepackage{ragged2e}
\usepackage{float}
\usepackage{pifont}
\newcommand{\cmark}{\checkmark}
\newcommand{\xmark}{\ding{55}}
\onlineid{0}
\vgtccategory{Research}

\title{EgoPoseVR: Spatiotemporal Multi-Modal Reasoning for Egocentric Full-Body Pose in Virtual Reality}

\author{
  \authororcid{Haojie Cheng}{0000-0002-9885-763X}, \authororcid{Shaun Jing Heng Ong}{0009-0005-7430-8467}, \authororcid{Shaoyu Cai}{0000-0001-8808-3442}, \\ 
  \authororcid{Aiden Tat Yang Koh}{0009-0007-5677-7045}, \authororcid{Fuxi Ouyang}{0009-0008-8618-8070}, and \authororcid{Eng Tat Khoo$^{*}$}{0000-0003-1295-3506} 
}

\authorfooter{
  \item
  	Haojie Cheng, Shaun Jing Heng Ong, Shaoyu Cai, Aiden Tat Yang Koh, and Eng Tat Khoo are with National University of Singapore.
  	E-mail: hjcheng@nus.edu.sg, jong1.05@nus.edu.sg, shaoyucai@nus.edu.sg, aiden@nus.edu.sg, etkhoo@nus.edu.sg.
   \item
   Fuxi Ouyang is with Singapore University of Technology and Design.
   E-mail: fuxi\_ouyang@mymail.sutd.edu.sg.
   \item
   $^{*}$ Corresponding author.
}

\teaser{
\vspace{0.13in}
  \centering
    \includegraphics[width=0.239\linewidth]{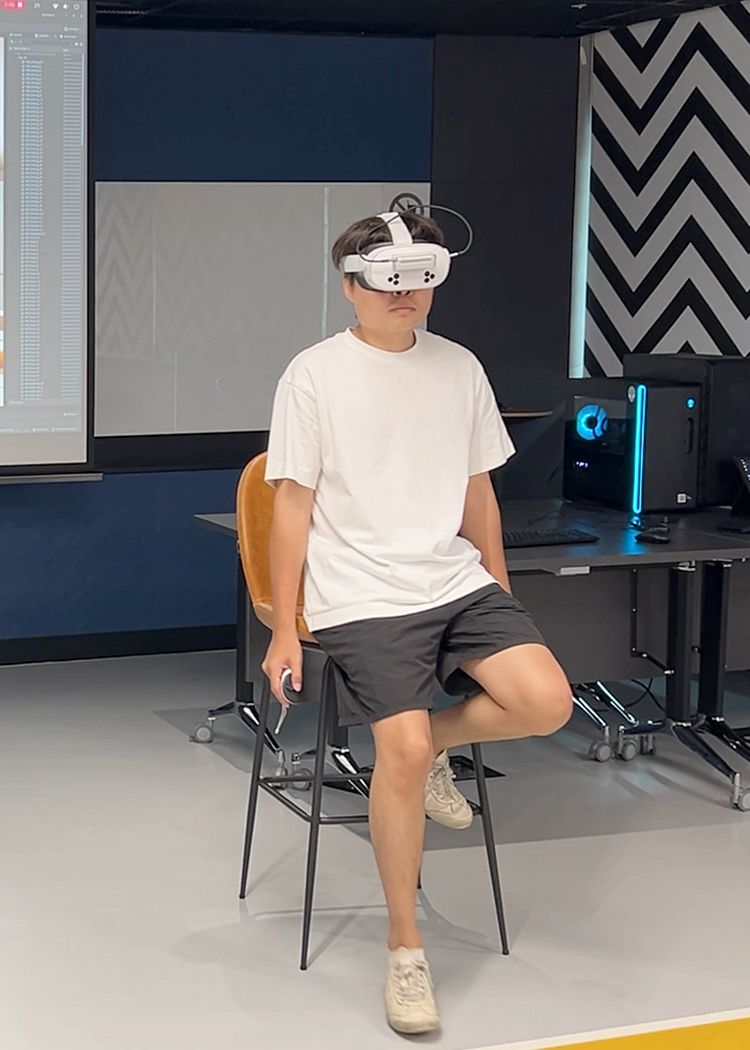}
    \includegraphics[width=0.239\linewidth]{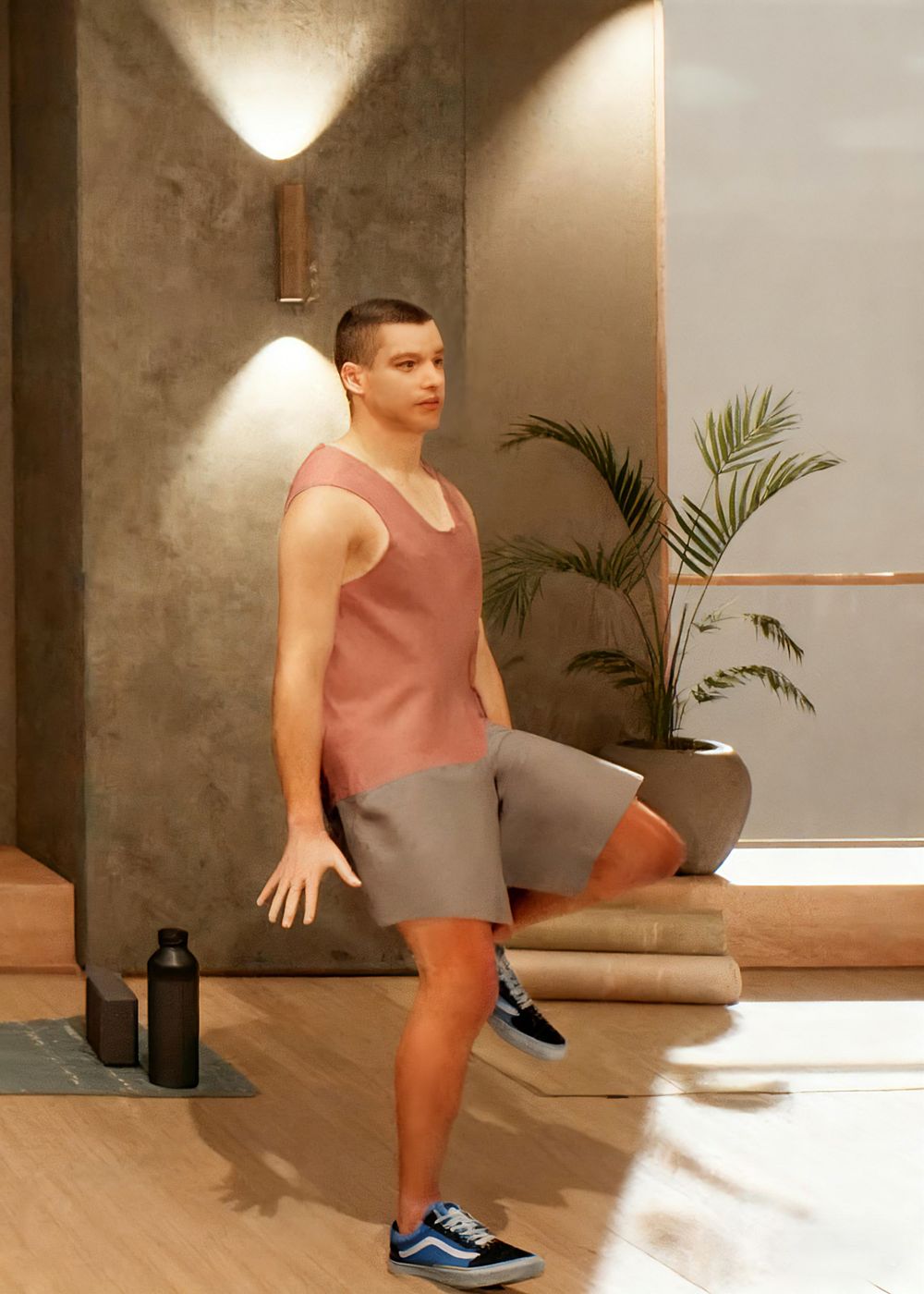}
 \hspace{0.0001in}
    \includegraphics[width=0.239\linewidth]{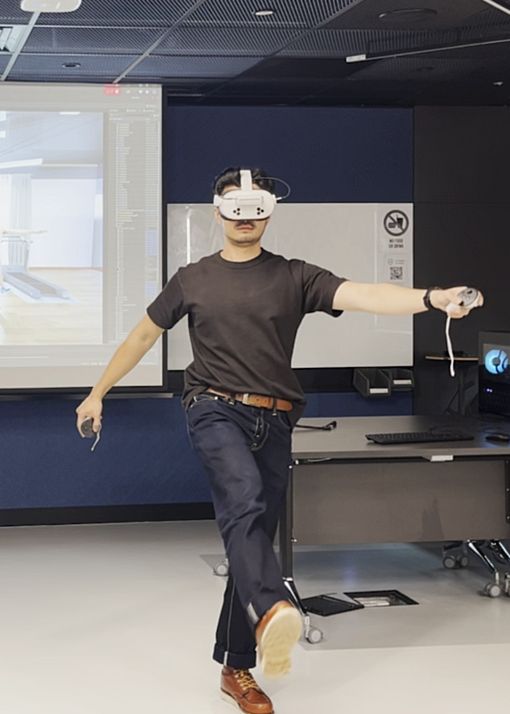}
    \includegraphics[width=0.239\linewidth]{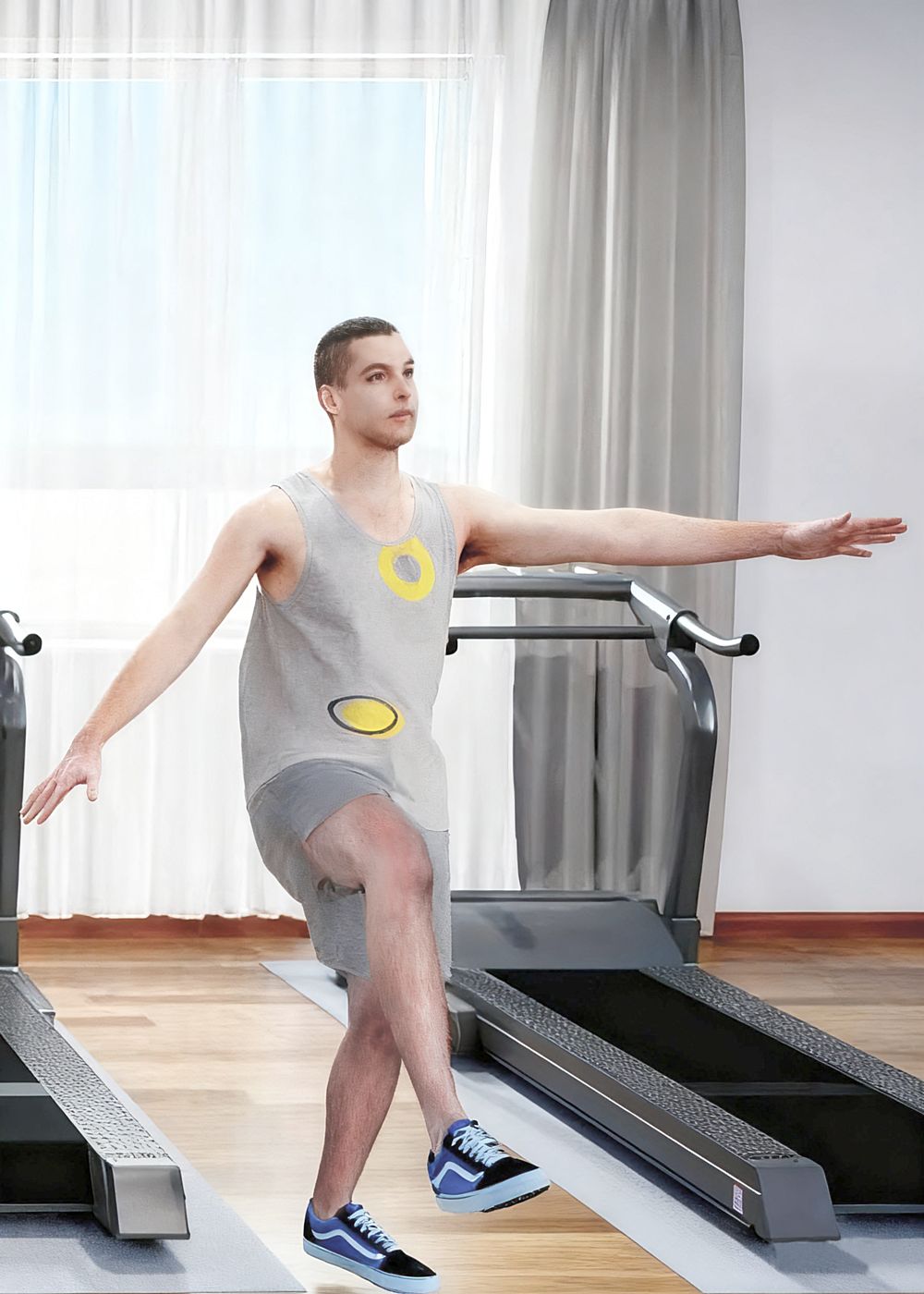}
    
\caption{Egocentric full-body pose estimation in VR enabled by an integrated setup that combines a VR headset with hand controllers and a headset-mounted downward-facing RGB-D camera. The configuration supports real-time and accurate tracking of both upper- and lower-body movements to facilitate natural avatar control and immersive interaction.}
  \label{fig:teaser}
}

\abstract{
    Immersive virtual reality (VR) applications demand accurate, temporally coherent full-body pose tracking. Recent head-mounted camera-based approaches show promise in egocentric pose estimation, but encounter challenges when applied to VR head-mounted displays (HMDs), including temporal instability, inaccurate lower-body estimation, and the lack of real-time performance. To address these limitations, we present \textit{EgoPoseVR}, an end-to-end framework for accurate egocentric full-body pose estimation in VR that integrates headset motion cues with egocentric RGB-D observations through a dual-modality fusion pipeline. A spatiotemporal encoder extracts frame- and joint-level representations, which are fused via cross-attention to fully exploit complementary motion cues across modalities. A kinematic optimization module then imposes constraints from HMD signals, enhancing the accuracy and stability of pose estimation. To facilitate training and evaluation, we introduce a large-scale synthetic dataset of over $1.8$ million temporally aligned HMD and RGB-D frames across diverse VR scenarios. Experimental results show that \textit{EgoPoseVR} outperforms state-of-the-art egocentric pose estimation models. A user study in real-world scenes further shows that \textit{EgoPoseVR} achieved significantly higher subjective ratings in accuracy, stability, embodiment, and intention for future use compared to baseline methods. These results show that \textit{EgoPoseVR} enables robust full-body pose tracking, offering a practical solution for accurate VR embodiment without requiring additional body-worn sensors or room-scale tracking systems.
} 


\begin{document}

\maketitle

\section{Introduction}
\label{sec:introduction}

Virtual reality (VR) enables a wide range of immersive applications, including embodiment entertainment \cite{starke2024categorical,do2024stepping,tao2023embodying}, physical training and rehabilitation \cite{liu2024facilitating, uhl2023tangible, cheng2023realistic}, and social collaboration \cite{wang2024pepperpose, rhee2020augmented,lammert2024immersive}. Many of these scenarios require accurate and spatiotemporally consistent full-body pose estimation to faithfully reflect the user's physical state and enable intuitive interaction. However, current consumer VR devices offer limited full-body tracking capabilities. Therefore, researchers have widely explored full-body pose estimation by integrating external systems into VR setups, such as optical marker-based motion capture systems \cite{eom2022neurolens, rahimian2016optimal} and third-person view cameras \cite{sun2024repose, zhu2023motionbert}. Although current optical marker-based systems can achieve high-accuracy pose tracking, they require large capture spaces and meticulous calibration \cite{van2018accuracy}. While third-person vision offers complete whole-body observations with minimal occlusion, its reliance on external viewpoints makes it unsuitable for immersive first-person applications such as VR, which inherently depend on egocentric sensing \cite{azam2024survey}.

\begin{figure*}[h]
\centering
\hspace{-0.10in}
\includegraphics[width=1.0\textwidth]{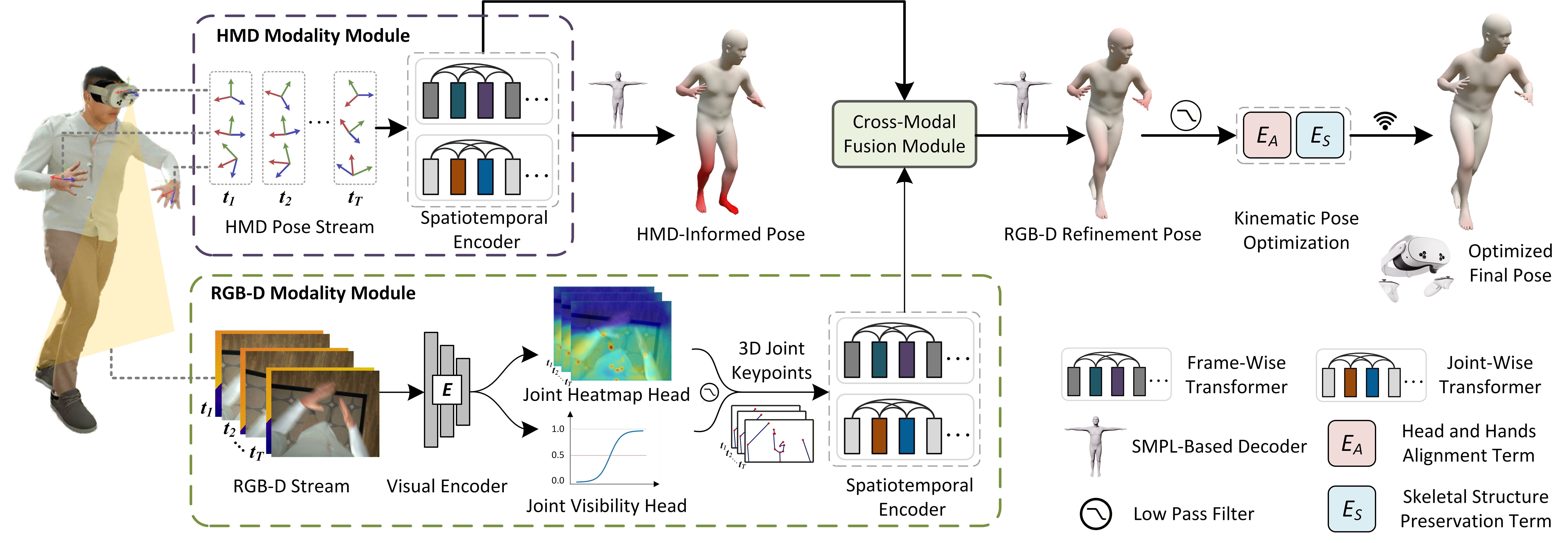}
\caption{Workflow of our EgoPoseVR framework for egocentric full-body pose estimation in VR. HMD motion and egocentric downward-facing RGB-D inputs are jointly encoded by the spatiotemporal module to predict full-body poses. A tailored kinematic optimization via energy functions ensures structurally consistent and immersive avatar rendering.}
\label{fig:pipeline}
\end{figure*}

Modern VR hardware, including head-mounted displays (HMDs) with hand-held controllers, provides stable and low-latency motion measurements for the user’s head and hands \cite{rojo2022accuracy}. However, inferring a full-body pose from only head and hand tracking is fundamentally ambiguous, as similar upper-body motions may correspond to multiple plausible full-body configurations \cite{jiang2022avatarposer, zheng2023realistic}. This limitation underscores the need to capture lower-body movement data in VR. One approach is to attach additional body-worn sensors on the lower limbs to provide extra motion cues ~\cite{zheng2018hybridfusion, huang2018deep,von2017sparse,yi2021transpose,yi2022physical,dai2024hmd, ponton2023sparseposer}. Among these, inertial measurement units (IMUs) are the most widely used, as they are compact and robust to occlusion, but require precise calibration and suffer from drift and usability burdens ~\cite{ponton2023sparseposer}. Another promising alternative is to use an egocentric downward-facing camera~\cite{rhodin2016egocap}, which captures first-person observations of the user’s body. Recent studies in computer vision have demonstrated that this configuration substantially improves the accuracy of lower-body pose \cite{xu2019mo,tome2019xr,kang2023ego3dpose,yang2024egoposeformer,lee2025rewind,kang2024attention,millerdurai2024eventego3d,luo2024real, cuevas2024simpleego}, demonstrating improved accuracy of lower-body pose estimation. However, the direct transfer of these methods to VR is impeded not only by camera access constraints \cite{fan2025emhi}, but also by fundamental challenges in modality integration.

Integrating HMD motion data with egocentric visual input in a unified framework poses significant challenges. HMD motion data are spatially sparse, structured, and low-dimensional, whereas egocentric vision data is high-dimensional and highly sensitive to even minor head movements, which often induce abrupt scene changes or motion blur and compromise temporal stability \cite{kang2023ego3dpose,wang2021estimating,azam2024survey}. Moreover, existing egocentric vision methods are already computationally demanding \cite{azam2024survey}, and extending them with additional VR motion signals from head-mounted displays and controllers further increases computational complexity and inference latency. Beyond these representational differences, progress is further constrained by the absence of multi-modal datasets that provide temporally synchronized headset motion and egocentric vision data in VR environments. These challenges are further amplified in VR, where users are highly sensitive to inaccuracies and latency \cite{roth2020construction}, motivating the development of a principled framework that unifies motion and vision under practical immersive conditions.


To address these issues, we present \textit{EgoPoseVR} as shown in Fig. \ref{fig:pipeline}, a dual-stream spatiotemporal framework for egocentric full-body pose estimation that combines HMD-based motion data with visual input from a downward-facing egocentric camera, specifically designed for VR scenarios. Each stream is encoded with spatiotemporal context, capturing both temporal dynamics and joint-level dependencies. Within the RGB-D modality, we design a visibility-aware temporal joint detection module that mitigates instability from motion blur and scene variations, thereby enhancing robustness to visual noise. To exploit complementary information across modalities, we further incorporate a cross-attention fusion module that refines HMD motion features with visual cues. Finally, a kinematic optimization module is introduced to improve pose accuracy by enforcing consistency with VR signals while preserving skeletal structure. To support both the training and inference, we construct a multi-modal dataset tailored for egocentric full-body pose estimation in VR. Evaluations on the synthetic dataset demonstrate that \textit{EgoPoseVR} achieves more accurate and stable pose estimation than existing state-of-the-art methods, while maintaining real-time performance at $97$ FPS. A real-world user study further validated these findings, showing significantly higher subjective ratings in accuracy, stability, embodiment, and future-use intention.

In summary, the main contributions of our work are as follows:

\begin{itemize}
    \item We introduce \textit{EgoPoseVR}, a novel egocentric pose estimation system for VR that combines HMD motion signals and egocentric visual observations through a headset-mounted RGB-D camera, enabling deployable full-body tracking without external infrastructure.

    \item We propose a dual-stream spatiotemporal architecture that encodes frame-wise dynamics and joint-level dependencies, integrates complementary cues through a cross-modal fusion module and refines predictions with a kinematic optimizer enforcing structural consistency with VR-tracked signals.
    
    \item We construct the first large-scale synthetic dataset for egocentric full-body pose estimation in VR, featuring temporally synchronized HMD motion and RGB-D data with $2$D and $3$D joint annotations, comprising $1.8$ million frames across varied VR scenes and applicable to different HMD–camera setups.
\end{itemize}

\begin{table*}[t]
\small
\centering
\caption{Comparison of representative datasets for egocentric full-body human pose estimation.}
\vspace{-0.05in}
\renewcommand{\arraystretch}{0.9}
\setlength\tabcolsep{3pt}
\begin{tabular}{p{25mm}|C{19mm} C{19mm} C{19mm} C{17mm} C{17mm} C{17mm} C{22mm}}
\hlineB{3}
\centering  & Mo2Cap2 \cite{xu2019mo} & xR-EgoPose \cite{tome2019xr} & EgoGlass \cite{zhao2021egoglass} & UnrealEgo \cite{akada2022unrealego} & ARES \cite{li2023ego} & SynthEgo \cite{cuevas2024simpleego}& EgoPoseVR (Ours) \\
\hline
\hline
Camera Lens Type &  Fisheye& Fisheye& Perspective &  Fisheye&  Perspective & Perspective& Perspective  \\
Mono/Stereo & Mono & Mono &  Stereo& Stereo & Mono &  Stereo & Mono   \\
Body Visibility &  \cmark&  \cmark&  \cmark&  \cmark&  \xmark &  \cmark& \cmark  \\
Environment & In- \& Outdoor & Mostly Indoor& Indoor&  Indoor&  In- \& Outdoor&  In- \& Outdoor& In- \& Outdoor   \\
Joint Location & \cmark & \cmark & \cmark & \cmark & \cmark & \cmark & \cmark \\
Joint Rotation & \xmark  & \xmark  & \xmark & \xmark  & \cmark & \cmark & \cmark  \\
Depth Availability & \xmark & \xmark & \xmark & \xmark &  \xmark& \xmark & \cmark\\
HMD Adaptability & \xmark & \xmark & \xmark & \xmark &  \xmark& \cmark (Implicit) & \cmark (Explicit)\\
Temporal Continuity & \xmark & \xmark & \xmark & \xmark &  \xmark& \xmark & \cmark\\
Dataset Size& $530k$ & $383k$ & $2\times170k$ & $2\times450k$ & $1.2M$ & $2\times60k$ & $1.8M$   \\
\hlineB{3}
\end{tabular}
\label{tab:datasetcomparison}
\end{table*}

\section{Related Works}
\label{sec:relatedworks}

\subsection{Egocentric Pose Estimation in VR}
\label{subsec: egoposeworks}

In VR applications, a common and convenient configuration is to estimate full-body pose using only HMD input, without relying on additional sensors or visual observations. Physics-based approaches like QuestSim \cite{winkler2022questsim} and QuestEnvSim \cite{lee2023questenvsim} generate physically plausible motion via reinforcement learning and simulation, but are difficult to integrate into real-time systems due to their reliance on non-differentiable simulators. In contrast, data-driven models (e.g., AvatarPoser \cite{jiang2022avatarposer}, AvatarJLM \cite{zheng2023realistic} and EgoPoser \cite{jiang2024egoposer}) regress poses directly using neural architectures with temporal modeling or inter-joint constraints, which contribute to improved performance of pose estimation. More recently, diffusion-based methods such as AGRoL \cite{du2023avatars} and EgoEgo \cite{li2023ego} produce realistic motion sequences, though their iterative sampling and prediction of future frames limit real-time applicability. These HMD-only methods often struggle to predict lower-body pose accurately, due to the lack of global body context in spatially sparse HMD input.

To enhance full-body pose estimation beyond spatially sparse HMD inputs, recent work has explored integrating additional body-worn sensors (e.g., IMUs) \cite{fan2025emhi}. To recover full-body motion from sparse signals, DIP \cite{huang2018deep} employs recurrent sequence modeling, while SIP \cite{von2017sparse} frames the task as an optimization over sparse-to-dense pose trajectories. Other methods incorporate biomechanical constraints and structural priors to improve physical plausibility, such as TransPose \cite{yi2021transpose} and PIP \cite{yi2022physical}. Real-time applications have also been explored in HMD-Poser \cite{dai2024hmd}, which combines lightweight neural networks with online shape adaptation for practical VR use. Beyond IMUs, alternative sensing strategies include wearable cameras attached to the torso or limbs, as in Nymeria \cite{ma2024nymeria} and EgoSim \cite{hollidt2024egosim}, which combine RGB camera with motion capture suits. However, these wearable sensors typically require bespoke hardware setup and are vulnerable to magnetic interference \cite{kaufmann2021pose}, limiting their scalability and practicality in VR contexts.

\vspace{-0.05in}
\subsection{Motion Tracking from Egocentric Camera}
\label{subsec:sparsemotion}

A prominent direction in egocentric pose estimation refines lower-body predictions by employing a downward-facing head-mounted camera to capture visual observations \cite{azam2024survey}. Early methods such as Mo2Cap2 \cite{xu2019mo}, xR-EgoPose \cite{tome2019xr} and EgoPW \cite{wang2022estimating} demonstrate feasibility but are fundamentally limited by severe self-occlusions and the lack of explicit kinematic modeling. Recent approaches extend beyond monocular baselines by incorporating additional cues. Geometry-aware methods, such as Ego3DPose \cite{kang2023ego3dpose}, exploit stereo correspondence and orientation cues to mitigate self-occlusions. Scene-aware methods, exemplified by Scene-Aware EgoPose \cite{wang2023scene}, integrate depth and voxelized context to enforce plausibility. More recently, generative motion models, including REWIND \cite{lee2025rewind}, adopt diffusion-based formulations for real-time whole-body estimation. EgoPoseFormer \cite{yang2024egoposeformer} and Ego4View \cite{akada2025bring} propose coarse-to-fine refinement strategies to progressively lift egocentric observations to 3D poses. While improving robustness, such coarse-to-fine designs incur high inference latency and rely heavily on pelvis-to-camera constraints.

Current image-based egocentric pose estimation methods are primarily evaluated on synthetic datasets (Table~\ref{tab:datasetcomparison}), such datasets are widely adopted as they enable scalable data generation without site-specific constraints, support diverse and controllable variations in environments and clothing, and avoid the need for additional complex motion capture setups. While synthetic data may not fully capture all aspects of real-world variability, it provides a practical and controllable foundation for training and evaluation. Rather than addressing the sim-to-real gap solely through expanded data coverage, we focus on algorithmic designs that improve robustness under distribution shifts between synthetic and real-world scenarios. Beyond this limitation, current methods typically perform pose estimation on a per-frame basis, which leads to temporal inconsistency. Self-occlusion and motion blur from abrupt head movements can further distort the egocentric view and destabilize predictions. In addition, prior works have overlooked the practical constraint that commercial headsets (e.g., Meta Quest and Apple Vision Pro) do not grant the data access of downward-facing fisheye cameras, which restricts their applicability and prevents rigorous validation in real-world VR environments \cite{luo2024real}.


\subsection{Spatiotemporal Modeling for Temporal Dynamics}
\label{subsec:temporalmodeling}

Spatiotemporal modeling is already widely used in third-person 3D human pose estimation, as temporal context alleviates frame-wise ambiguities and enforces kinematic consistency across frames \cite{dosovitskiy2020image}. Transformer-based architectures such as PoseFormer \cite{zheng20213d}, MixSTE \cite{zhang2022mixste}, and MHFormer \cite{li2022mhformer} leverage token-wise attention to capture long-range dependencies across RGB sequences, while diffusion-based models, like DiffPose \cite{feng2023diffpose} and S$^{2}$Fusion \cite{tang2024unified}, further incorporate probabilistic temporal priors. Despite their effectiveness, these methods generally use dense visual frames and are often tailored for offline settings. Recent efficient designs such as UNSPAT \cite{lee2024unspat} and HoT \cite{li2024hourglass} attempt to reduce the computation of full-frame processing by aligning spatial positions or pruning. In contrast, several studies demonstrate that reliable temporal modeling can be achieved without densely sampled video image sequences. PhaseMP \cite{shi2023phasemp} and ReMP \cite{jang2025remp} encode long-range dynamics from sparse motion streams using phase representations or reusable motion priors, while KTPFormer \cite{peng2024ktpformer} improves spatial-temporal stability by injecting kinematic priors.

Despite advances in leveraging spatiotemporal visual information, egocentric pose estimation remains hindered by dataset limitations. As shown in Table~\ref{tab:datasetcomparison}, popular benchmarks such as Mo2Cap2 \cite{xu2019mo}, xR-EgoPose \cite{tome2019xr} and UnrealEgo \cite{akada2022unrealego} primarily provide isolated frames without temporally aligned image sequences, restricting the development of spatiotemporal models. Moreover, most datasets do not include explicit annotations of joint rotations (e.g., EgoGlass \cite{zhao2021egoglass}), which are required to simulate dynamic six degrees of freedom (6DoF) motion signals from headsets and controllers. The absence of both temporal dynamics and rotation information further prevents consistent alignment between headset and controller signals and egocentric visual inputs. This gap motivates our creation of a spatiotemporally aligned egocentric pose dataset tailored for VR applications.


\section{Methodology}

\subsection{Overview}
\label{subsec:Overview}

We tackle the task of real-time full-body $3$D pose estimation for VR, predicting both joint orientations and positions from synchronized egocentric signals. The input consists of temporal motion data from the headset and hand controllers, complemented by downward-facing RGB-D observations captured by a commercial camera mounted on the headset. In this setting, ensuring computational efficiency while enabling complementary use of modalities without cross-interference is a central challenge of our design.

To this end, we propose \textit{EgoPoseVR}, a dual-stream spatiotemporal framework (Fig.~\ref{fig:pipeline}). Each stream is processed by a designed spatiotemporal encoder built with Transformer layers that capture frame-wise temporal dynamics and joint-wise dependencies, yielding modality-specific representations. The motion stream operates on headset and controller trajectories, while the visual stream encodes temporally stable joints extracted from RGB-D input, providing compact visual-geometric cues. The two representations are fused in a shared latent space to predict a refined full-body pose, which is subsequently refined by a tailored kinematic optimizer to ensure consistency with VR tracking signals and skeletal structural constraints. The final output is a temporally stable and spatially accurate full-body pose estimation, well-suited for low-latency VR deployment.

\subsection{Input Modalities and Pose Representation}
\label{subsec:inputoutput}

The input signal $\mathbf{x}_t$ at the current frame $t$ from the HMD consists of spatially sparse motion measurements. For each device $i \in \{H, C_L, C_R\}$ corresponding to the head, left controller and right controller respectively, we extract the global 3D position $\mathbf{p}^i_t \in \mathbb{R}^3$, 6D orientation $\boldsymbol{\theta}^i_t \in \mathbb{R}^6$, linear velocity $\mathbf{v}^i_t \in \mathbb{R}^3$, and 6D angular velocity $\boldsymbol{\omega}^i_t \in \mathbb{R}^6$. These components are concatenated into a global motion descriptor $\mathbf{g}^i_t = [\mathbf{p}^i_t,\ \boldsymbol{\theta}^i_t,\ \mathbf{v}^i_t,\ \boldsymbol{\omega}^i_t] \in \mathbb{R}^{18}$. To reduce sensitivity to global pose and facilitate body-centric reasoning, we explicitly compute the relative pose of each controller with respect to the headset. For each controller $c \in \{C_L, C_R\}$, we define the head-relative position $\tilde{\mathbf{p}}^c_t \in \mathbb{R}^3$ and 6D rotation $\tilde{\boldsymbol{\theta}}^c_t \in \mathbb{R}^6$ by transforming the controller pose into the HMD coordinate frame. These features are concatenated into local motion descriptor $\mathbf{r}^c_t = [\tilde{\mathbf{p}}^c_t,\ \tilde{\boldsymbol{\theta}}^c_t] \in \mathbb{R}^{9}$. The HMD input vector $\mathbf{x}_t \in \mathbb{R}^{72}$ is constructed by concatenating all motion descriptors and can be written as:

\begin{equation}
    \mathbf{x}_t = [\mathbf{g}^H_t,\ \mathbf{g}^{C_L}_t,\ \mathbf{g}^{C_R}_t,\ \mathbf{r}^{C_L}_t,\ \mathbf{r}^{C_R}_t] \in \mathbb{R}^{72}.
\label{Eq. HMDinput}
\end{equation}

In parallel, a downward-facing RGB-D camera mounted on the HMD captures egocentric visual observations, denoted as $\mathbf{y}_t \in \mathbb{R}^{C \times H \times W}$, where $C$, $H$ and $W$ denote the number of channels, height and width, respectively. To leverage the temporal continuity of HMD motion and RGB-D image data, we construct a preset window of $T$ consecutive frames ending at time $t$, i.e., $\mathbf{X} = [\mathbf{x}_{t-T+1}, \dots, \mathbf{x}_{t}] \in \mathbb{R}^{T \times 72}$, $\mathbf{Y} = [\mathbf{y}_{t-T+1}, \dots, \mathbf{y}_{t}] \in \mathbb{R}^{T \times C \times H \times W}$. Therefore, the final input for each frame $t$ is defined as the combination of HMD motion and visual modalities:

\begin{equation}
    \mathcal{I}_{t-T+1:t} = \{\mathbf{X},\ \mathbf{Y}\}.
\label{Eq. HMDplusRGBDinput}
\end{equation}

Our model predicts full-body human pose using the SMPL parameterization \cite{loper2023smpl}. Following prior work~\cite{jiang2022avatarposer}, we focus on the main skeletal structure and exclude the two palm joints, resulting in $22$ output joints. For each joint, the model predicts its rotation relative to its parent joint in the kinematic chain. This yields an output of dimension $\mathbb{R}^{132}$ per frame, corresponding to $22$ joints represented by a $6$D rotation for SMPL-based avatar control.

\subsection{HMD and RGB-D Feature Modeling}
\label{subsec:DLModel}

\subsubsection{HMD Stream Encoder}
To effectively capture temporal dynamics and spatial structure from egocentric motion data, we introduce a spatiotemporal encoder that operates on HMD-based measurements, as defined in Eq. (\ref{Eq. MotionEq}). Given the temporal HMD descriptors $\mathbf{X}$ in Eq. (\ref{Eq. HMDplusRGBDinput}), we first project each frame into a latent space via a linear embedding $E_{l}(\cdot)$, producing a sequence of feature vectors. To capture long-range temporal dependencies and mitigate the limitations of frame-wise or short-term modeling, we adopt a frame-wise transformer encoder $\mathcal{T}_{f}(\cdot)$. Its self-attention mechanism enables dynamic weighting of past motion cues across the entire temporal window, thereby facilitating robust trend extraction and improved temporal coherence in downstream pose estimation.

\begin{equation}
\mathbf{M}_{t} = \mathcal{T}_j \left( \mathbf{F} \left( \left[ \mathcal{T}_f \left( E_{l}(\mathbf{X}) \right) \right]_t \right) + \mathbf{e}_j \right).
\label{Eq. MotionEq}
\end{equation}


To retain temporal context while focusing on the current $t$-th frame, we extract the final temporal token and project it into a structured pose representation via a multi-layer perceptron (MLP), denoted as $\mathbf{F}(\cdot)$. This step summarizes long-term dynamics into a frame-specific feature, ensuring that subsequent modules operate with temporally aware information. To encourage the model to learn joint-specific semantics rather than treating all joints uniformly, we add a learnable embedding $\mathbf{e}_j$ to each joint feature. This provides an identity cue that helps the network disambiguate joints with similar motion patterns and facilitates reasoning about cross-joint relationships. Finally, we further refine the joint features using a joint-wise transformer encoder $\mathcal{T}_{j}(\cdot)$. This allows the network to model kinematic constraints and generate anatomically consistent pose hypotheses. The resulting joint-wise feature representation $\mathbf{M}_{t}$ forms a tensor of shape $\mathbb{R}^{J \times S}$, where $J$ is the number of joints and $S$ is the per-joint feature dimension.

\subsubsection{RGB-D Stream Encoder.}
\label{subsubsec:visualstream}

While the HMD stream captures head and hand trajectories and can coarsely infer lower-body motion, its predictions remain ambiguous. The same upper-body movement can correspond to multiple plausible lower-body poses. To resolve this ambiguity, we introduce a complementary vision stream that processes RGB-D inputs from a downward-facing camera. 

{
\renewcommand{\baselinestretch}{0.9}\selectfont
\small
\begin{algorithm}
\small
\SetKwInOut{Input}{Input}
\SetKwInOut{Output}{Output}
\Input{$\mathbf{Y}$, $\mathbf{Z}$, $T$, $J$ and $\zeta$}
\Output{Refined 3D joint keypoint sequence $\tilde{\mathbf{Z}}$}

\uIf{$\mathbf{Y}$ is the first joint keypoint batch}{
        \ForEach{$(t',j)$ where $t' \in [t-T+1, t]$, $j \in [1, J]$}{
            $\tilde{\zeta}_{t',j} \leftarrow \mathcal{F}(\zeta_{t',j})$\;
        }
    Joint-wise mask $\eta \leftarrow \text{ReLU}(\tilde{\zeta} - 0.5)$,\quad $\tilde{\mathbf{Z}} \leftarrow \mathbf{Z} \odot \eta$\;
}
\Else{
    Let $N$ be the number of newly appended frames\;
    New frames $\mathcal{N}$: $[t\!-\!N\!+\!1,\ t]$, Previous frames $\mathcal{P}$: $[t\!-\!T\!+\!1,\ t\!-\!N]$ \;
    \If{$N > 0$}{
        $\mathbf{Z}^{+},\ \zeta^{+} \leftarrow \mathbf{Z}_{\mathcal{N}},\ \zeta_{\mathcal{N}}$, \ $\mathbf{Z}^{-},\ \zeta^{-} \leftarrow \tilde{\mathbf{Z}}_{\mathcal{P}},\ \tilde{\zeta}_{\mathcal{P}}$\;
            
        $\hat{\zeta} =\text{Concat}(\zeta^{-}, \zeta^{+})$\;
    
        \ForEach{$(t,j)$ where $t' \in \mathcal{N}$, $j \in [1, J]$}{
            $\tilde{\zeta}_{t',j} \leftarrow \mathcal{F}(\hat{\zeta}_{t',j})$\;
            }
        $\eta \leftarrow \text{ReLU}(\tilde{\zeta}_{\nu} - 0.5)$\;      $\tilde{\mathbf{Z}}^{+} \leftarrow \mathbf{Z}^{+} \odot \eta$,\quad $\tilde{\mathbf{Z}} \leftarrow \text{Concat}(\mathbf{Z}^{-}, \tilde{\mathbf{Z}}^{+})$\;
    }
}
\caption{Temporal refinement of 3D joint estimates}
\label{algo:jointoptimization}
\end{algorithm}
}

To utilize the RGB-D stream effectively, a convolutional heatmap encoder, i.e., a ResNet backbone with a feature pyramid network (FPN) \cite{lin2017feature}, first processes the RGB-D input to estimate $3$D joint keypoint sequence $\mathbf{Z} \in \mathbb{R}^{T \times J \times 3}$. Compared to directly apply spatiotemporal modeling on dense image sequences $\mathbf{Y}$ in Eq. \ref{Eq. HMDplusRGBDinput}, $3$D joint keypoints significantly reduce the computational complexity of temporal reasoning from image-level (i.e., $\mathcal{O}(T \cdot C \cdot H \cdot W)$) to joint-level (i.e., $\mathcal{O}(T \cdot J)$). In addition, the explicit $3$D joint supervision stabilizes training and provides spatially grounded intermediate features.

To improve the robustness and stability of 3D joint estimation, we further introduce a new refinement strategy, as outlined in Alg. \ref{algo:jointoptimization}. Because each heatmap always produces a peak, joints that are outside the camera’s view may be erroneously detected as visible, resulting in false positives. This misrepresentation can lead to inaccurate joint localization and unstable predictions in real-world scenes. To address this, we design a visibility-aware probability prediction module, which outputs an initial visibility probability sequence $\zeta \in \mathbb{R}^{T \times J}$ indicating the likelihood that each joint is visible in the current view. Since the visual encoder processes multiple frames jointly, a one Euro filter $\mathcal{F}$ is further added to smooth $\zeta$. In addition, to accelerate inference, we design an incremental feature update scheme that extracts features only from newly appended frames and fuses them with cached features from previous frames. This design significantly reduces redundant computation while preserving temporal context. 

Following the estimation of the refined 3D joint keypoint sequence $\tilde{\mathbf{Z}}$ calculated by Alg. \ref{algo:jointoptimization}, the same spatiotemporal encoder with the architecture as Eq. \ref{Eq. MotionEq} is applied to frame-wise and joint-wise keypoint features $\mathbf{N}_{t} \in \mathbb{R}^{J\times S}$ from $\tilde{\mathbf{Z}}$. These features capture the temporal dynamics and inter-joint relationships of image-derived keypoints and are compatible with the HMD stream for subsequent cross-attention fusion. Notably, although both streams employ the identical spatiotemporal encoder design, their parameters are trained independently.

\begin{equation}
\vspace{-0.05in}
\mathbf{N}_{t} = \mathcal{T}_j \left( \mathbf{F} \left( \left[ \mathcal{T}_f \left( E_{l}(\tilde{\mathbf{Z}}) \right) \right]_t \right) + \mathbf{e}_j \right).
\label{Eq. RGBDEq}
\end{equation}

\subsubsection{Cross-Modal Spatiotemporal Integration.}

To integrate complementary information from the RGB-D stream with the HMD stream, we employ a standard multi-head cross-attention mechanism \cite{vaswani2017attention}. The HMD features $\mathbf{M}_{t}$ serve as queries, while RGB-D features $\mathbf{N}_{t}$ provide the keys and values, enabling HMD features to selectively attend to relevant visual cues. The resulting enriched representation  $\tilde{\mathbf{M}}_{t} = \text{CrossAttn}(\mathbf{M}_{t}, \mathbf{N}_{t})$ denotes HMD features refined with visual guidance. 

The fused features $\tilde{\mathbf{M}}_{t}$ are then fed into two parallel feed-forward networks, $\mathcal{T}_{g} (\cdot)$ and $\mathcal{T}_{l} (\cdot)$, to decode joint rotations represented in a continuous 6D rotation format. The first branch, $\mathcal{T}_{g}$, focuses on the root joint (i.e., the pelvis), whose global rotation determines the overall body orientation. Given its relatively stable and smooth trajectory, the pelvis benefits from a dedicated modeling pathway. The second branch, $\mathcal{T}_{l}$, processes the local rotation of the remaining joints to capture finer articulation patterns. This architectural decoupling reduces interference across spatial scales and facilitates more precise attention allocation to relevant features.

The $6$D joint rotation representations are subsequently propagated through a forward kinematics (FK) layer with the VR headset’s world coordinates to derive the full-body $3$D joint positions. To mitigate high-frequency jitter and improve temporal coherence across frames, we further apply a one Euro filter $\mathcal{F}$ to the output position trajectories, yielding smoother full-body joint position sequences $\mathbf{P}_{t} \in \mathbb{R}^{22 \times 3}$ at the current $t$-th frame:

\begin{equation}
\mathbf{P}_t = \mathcal{F} \left( \text{FK} \left(\mathcal{T}_{g}( \tilde{\mathbf{M}}_{t}), \mathcal{T}_{l}( \tilde{\mathbf{M}}_{t})  \right) \right).
\label{Eq. cross_attn}
\end{equation}

\subsection{Kinematic Pose Optimization via Energy Functions}
\label{subsec:skeletonoptimization}

Despite using headset and controller poses in the system input, the network may still yield inaccurate joint positions due to occlusions, sensor noise, and the ambiguity of inferring full-body motion from only head and hand tracking. Inverse kinematics (IK) can improve temporal continuity but remains under-constrained with head- and hand-only inputs, producing ambiguous or implausible lower-body poses without motion priors \cite{winkler2022questsim}. To address this, we propose an energy-based kinematic skeleton optimization that refines joint positions while preserving skeletal consistency.

\begin{figure}[h]
\centering
\subfloat[Original Pose]{
  \hspace{-0.15in}
\includegraphics[width=0.16\textwidth]{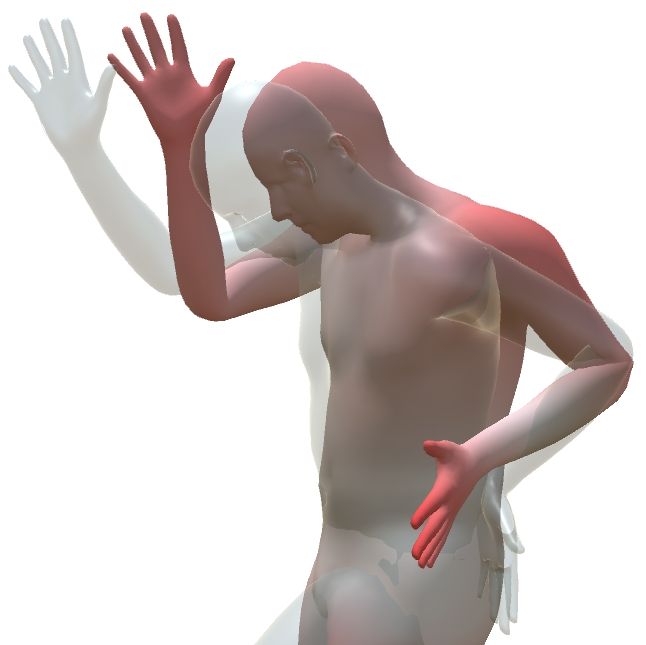}
}
\subfloat[$E_{\text{A}}$ Only]{
  \hspace{-0.05in}
\includegraphics[width=0.16\textwidth]{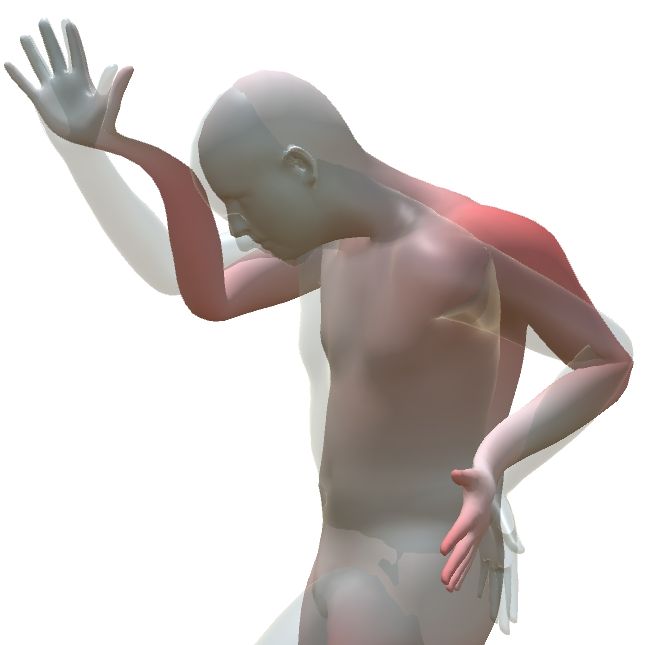}}
\subfloat[$E_{\text{A}}$ + $E_{\text{S}}$]{
  \hspace{-0.05in}
\includegraphics[width=0.16\textwidth]{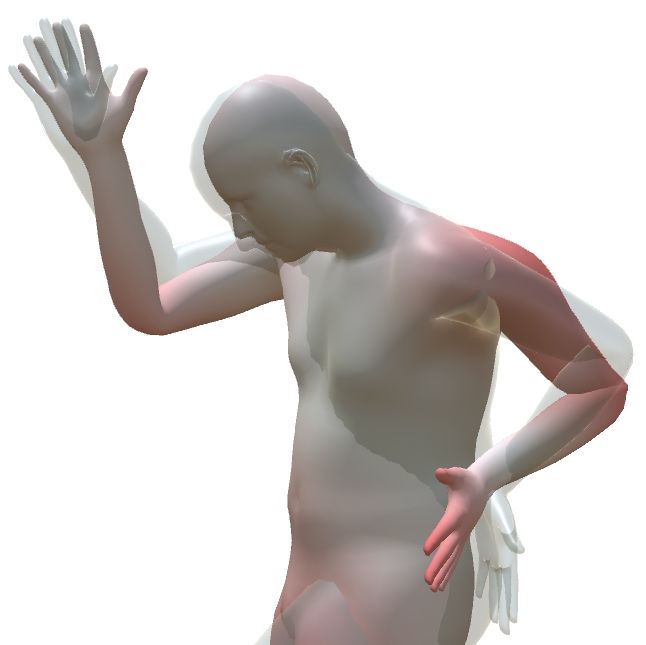}}

\vspace{-0.05in}
\caption{Pose refinement results under different kinematic pose optimization settings. The semi-transparent mesh is the ground-truth pose, while the predicted mesh is color-coded by the pose error (redder indicates higher error).}
\label{fig:UECoordff}
\vspace{-0.10in}
\end{figure}

\vspace{-0.05in}
\subsubsection{Head and Hands Alignment Term}

To ensure that the optimized skeleton remains consistent with external positional observations from the input HMD signals, we define a position alignment energy that enforces consistency between optimized joint positions and external observations from the HMD and controllers. Specifically, let $\mathcal{O}$ denote the set of observable joints (head, left hand, right hand) with ground-truth positions $\mathbf{q}_k \in \mathbb{R}^3$ provided by the VR system, where $k \in \mathcal{O}$. Let $\tilde{\mathbf{p}}_k$ denote the initial network prediction, and $\mathbf{p}_k$ denotes the optimized result. The energy function for head and hands alignment is formulated as:

\begin{equation}
E_{\text{A}} = \sum_{k \in \mathcal{O}} \lambda_a \left\| \mathbf{p}_k - \mathbf{q}_k \right\|^2 + \sum_{k \notin \mathcal{O}} \lambda_s \left\| \mathbf{p}_k - \tilde{\mathbf{p}}_k \right\|^2.
\label{eq:align}
\end{equation}

The first term encourages $\mathbf{p}_k$ to align with $\mathbf{q}_k$ for visible joints, ensuring consistency with real-world sensor input. The second term acts as a self-regularization constraint for unobserved joints ($k \notin \mathcal{O}$), guiding their optimized positions to remain close to their initial predictions $\tilde{\mathbf{p}}_k$. This formulation ensures that the optimization adjusts the original predictions only when supported by external observations, while preserving plausible pose structure in the absence of ground-truth data. The weight factors $\lambda_a$ and $\lambda_s$ control the relative importance of alignment with observed joints and regularization of unobserved joints, respectively.

\subsubsection{Skeletal Structure Preservation Term}

To ensure local skeletal structure is preserved during optimization, we define a structure-preserving energy that penalizes deviations in both bone length and bone direction between neighbouring joints. Let $\tilde{\mathbf{J}}_{ij}$ and $\mathbf{J}_{ij}$ denote the initial and optimized joint direction vector prediction from joint $j$ to $i$, respectively. The constraint energy function for local structure preserving is defined as:

\begin{equation}
\vspace{-0.10in}
E_{\text{S}} = \sum_i \sum_{j \in \mathcal{M}(i)} \left[
\lambda_l \left\| \ell(\mathbf{J}_{ij}) - \ell(\tilde{\mathbf{J}}_{ij}) \right\|^2 +
\lambda_d \left\| \mathbf{J}_{ij} - \tilde{\mathbf{J}}_{ij} \right\|^2
\right],
\label{eq:structure}
\vspace{0.10in}
\end{equation}

{\noindent}where $\mathcal{M}(i)$ denotes the set of joints that are directly connected to joint $i$ in the skeletal hierarchy, typically defined by the SMPL kinematic tree, $ \ell(\cdot)$ represents the distance calculation of adjacent joints. The first term enforces bone length consistency, while the second term encourages directional consistency of neighboring joints. The weight factors $\lambda_l$ and $\lambda_d$ control the relative importance of bone length and direction preservation, respectively.

\section{Synthetic Dataset Generation for VR Avatar}
\label{subsec:EgoPoseVRDataset}

To address the lack of datasets suitable for immersive VR applications that require spatiotemporally aligned pose data, we introduce \textit{EgoPoseVR}, a synthetic dataset featuring egocentric RGB-D imagery, HMD motion trajectories, and full-body $2$D/$3$D pose annotations, as listed in Table~\ref{tab:datasetcomparison}. EgoPoseVR comprises $18,235$ motion sequences and $1.8$ million frames rendered at $320 \times 256$ resolution and 60 FPS, establishing a large-scale benchmark for training and evaluating VR avatar pose estimation systems.

The dataset is constructed with five key design objectives. First, to ensure motion diversity, we curate $2,350$ sequences from AMASS~\cite{mahmood2019amass}, guided by category-level annotations from BABEL~\cite{punnakkal2021babel}, covering a broad range of daily and task-oriented movements. Second, to promote scene and illumination generalization \cite{cheng2022fast}, we include $8$ large-scale indoor $3$D environments together with $100$ HDR panoramic maps, providing both indoor- and outdoor-style lighting conditions. Third, to enhance appearance variability, the dataset incorporates $100$ clothing sets ($50$ male and $50$ female) and $20$ footwear assets, increasing visual realism in egocentric views. Fourth, to ensure cross-headset compatibility, we provide both pelvis-centered and camera-centered joint coordinates. The latter are obtained by applying forward kinematics from the pelvis to the head, followed by a pre-calculated relative pose between the head and egocentric camera, enabling accurate egocentric projections across arbitrary headset configurations. Finally, to support efficient and temporally consistent egocentric rendering, all RGB-D sequences are rendered at $320 \times 256$ resolution with alignment to HMD and hand controller trajectories. To account for partial observability in egocentric views, EgoPoseVR additionally provides per-joint visibility labels for each frame, encoded as binary indicators of whether a joint lies within the camera’s field of view. Further implementation details of EgoPoseVR dataset are provided in the supplementary material for reproducibility.

\begin{figure}[h]
\centering
\includegraphics[width=0.23\textwidth]{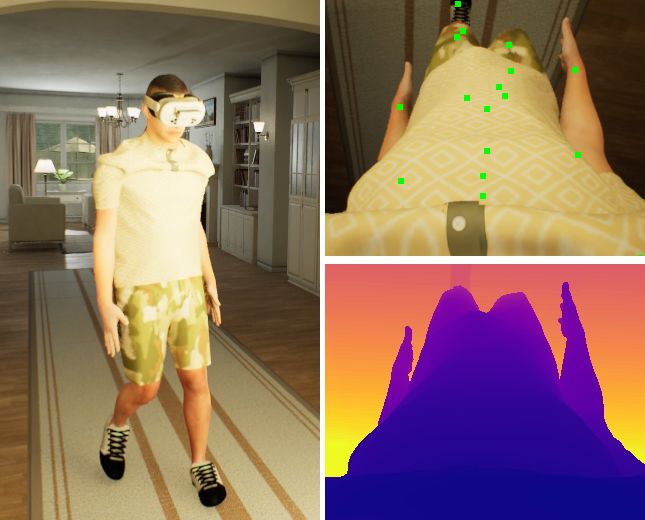} \hspace{0.01in}
\includegraphics[width=0.23\textwidth]{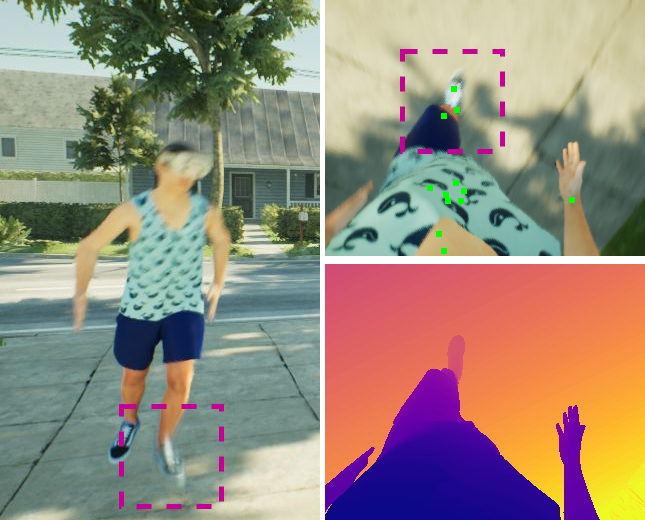}
\caption{Example images from our dataset showing third-person and egocentric RGB-D views. Green dots indicate 2D joint projections and pink dashed boxes mark the motion blur regions.}
\label{fig:datasetexample}
\vspace{-0.05in}
\end{figure}

\vspace{-0.05in}
\section{Cross-Device Data Transmission Pipeline}
\label{sec:DataTransmission}

To support real-time full-body pose estimation across separate sensing and rendering modules, we design a lightweight cross-device architecture based on a persistent WebSocket communication framework. The system consists of three components: an input client, a rendering client and an inference server. The input client integrates a VR HMD and an RGB-D camera, physically connected via USB Type-C. RGB-D images and HMD pose data are synchronized and transmitted in real time to the inference server via a persistent WebSocket connection. The inference server is responsible for running our EgoMotionVR model, and can host multiple models simultaneously, enabling flexible switching without altering the client architecture.



\begin{figure}[h]
\centering
\includegraphics[width=0.42\textwidth]{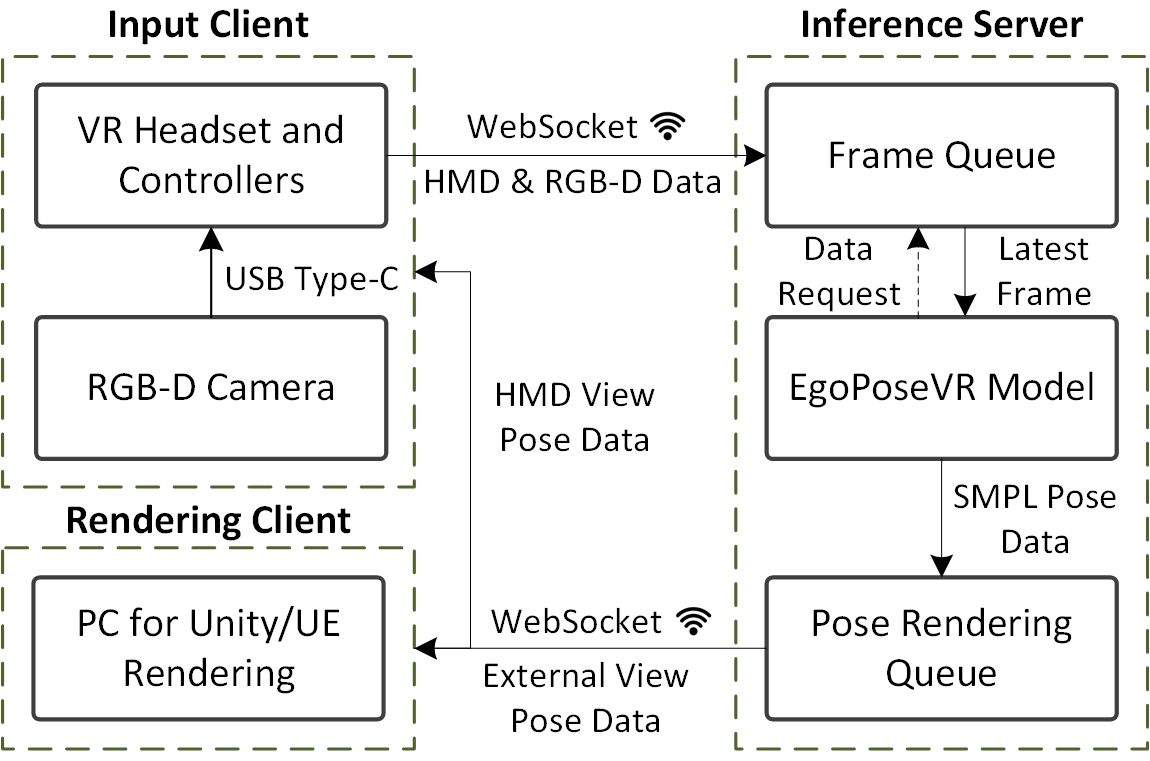}
\caption{System architecture illustrating cross-device data transmission for real-time SMPL pose estimation and visualization.}
\label{fig:datatransmission}
\end{figure}

Upon receiving data, the inference server temporarily stores the latest RGB-D frame in a buffer queue. Our EgoPoseVR model continuously retrieves the most recent frame on demand, estimates the full-body pose in SMPL format, and performs internal pose refinement to reduce noise and improve temporal stability. The refined SMPL pose is returned via WebSocket to the input client for in-headset visualization and to the rendering client for third-person monitoring, supporting synchronized third-person perspective feedback via Unity or UE render. The entire system operates asynchronously and supports bidirectional communication between the clients and server, enabling low-latency streaming of sensory data and immediate return of model predictions.

\begin{table*}[t]
\small
\centering
\caption{Quantitative results of module-level comparison with state-of-the-art methods on our synthetic dataset.
Row colors indicate performance improvement from the previous row:
\textcolor{lightyellow1}{\textbf{Yellow}} (\textless10\%),
\textcolor{darkgreen}{\textbf{Light Green}} (10–20\%),
\textcolor{darkgreen1}{\textbf{Dark Green}} (\textgreater20\%).}
\vspace{-0.05in}
\renewcommand{\arraystretch}{1.0}
\setlength\tabcolsep{1pt}
\begin{tabular}{C{5mm} | p{45mm}| C{16mm} C{16mm} C{19mm} C{19mm} C{16mm} C{16mm} C{16mm} C{4mm}}
\hlineB{3}
\multicolumn{2}{c|}{Method} & MPJPE-U $\downarrow$ & MPJPE-L $\downarrow$ & PA-MPJPE-U $\downarrow$ & PA-MPJPE-L $\downarrow$ & MPJRE-U $\downarrow$ & MPJRE-L $\downarrow$ & FPS $\uparrow$ \\
\cline{1-2}
\hline
\hline
A & EgoPoser \cite{jiang2024egoposer} &
$5.61 \pm 3.76$ & $8.45 \pm 5.99$ &
$4.13 \pm 2.56$ & $7.36 \pm 5.08$ &
$15.48 \pm 6.24$ & $10.72\pm 4.89$ &
216 $\pm$ 16\\

B & EgoPoser \cite{jiang2024egoposer} + EgoPoseFormer \cite{yang2024egoposeformer} &
\cellcolor{lightgreen}$4.92 \pm 3.35$ & \cellcolor{darkgreen}$6.44 \pm 4.92$ &
\cellcolor{lightgreen}$3.66 \pm 2.41$ & \cellcolor{darkgreen}$5.34 \pm 3.94$ &
\cellcolor{lightyellow}$15.39 \pm 6.21$ & \cellcolor{lightyellow}$9.97 \pm 4.63$ &
$29 \pm 1$\\

C & Proposed HMD + EgoPoseFormer \cite{yang2024egoposeformer} &
\cellcolor{darkgreen}$3.60 \pm 2.80$ & \cellcolor{lightyellow}$5.82 \pm 4.88$ &
\cellcolor{darkgreen}$2.68 \pm 1.81$ & \cellcolor{lightyellow}$4.83 \pm 3.80$ &
\cellcolor{darkgreen}$8.59 \pm 4.06$ & \cellcolor{darkgreen}$7.73 \pm 4.24$ &
$29 \pm 2$\\

D & Proposed HMD + Proposed RGB (M)  &
\cellcolor{lightyellow}$3.57 \pm 2.71$ & \cellcolor{lightgreen}$5.23 \pm 4.86$ &
\cellcolor{lightyellow}$2.55 \pm 1.80$ & \cellcolor{lightgreen}$4.20 \pm 3.73$ &
\cellcolor{lightyellow}$8.33\pm 4.04$ & \cellcolor{lightyellow}$7.29 \pm 4.28$ &
$155 \pm 6$ \\

E & EgoPoseVR (HMD + RGB-D (M) + KPO) &
\cellcolor{darkgreen}\textbf{$1.66 \pm 1.05$} & \cellcolor{lightyellow}\textbf{$4.75 \pm 4.35$} &
\cellcolor{lightgreen}\textbf{$2.05 \pm 1.53$} & \cellcolor{lightgreen}\textbf{$3.63 \pm 3.31$} &
\cellcolor{lightyellow}\textbf{$8.21 \pm 3.96$} & \cellcolor{lightyellow}\textbf{$6.74 \pm 3.94$} &
$97 \pm 6$\\
\hlineB{3}
\end{tabular}
\label{tab:Quantitative_DifferentMethods}
\vspace{-0.05in}
\end{table*}

\vspace{-0.05in}
\begin{figure*}[t]
\centering
\subfloat[Leaping Motion\label{fig:Competitor:leaping}]{%
  \includegraphics[height=0.165\textwidth]{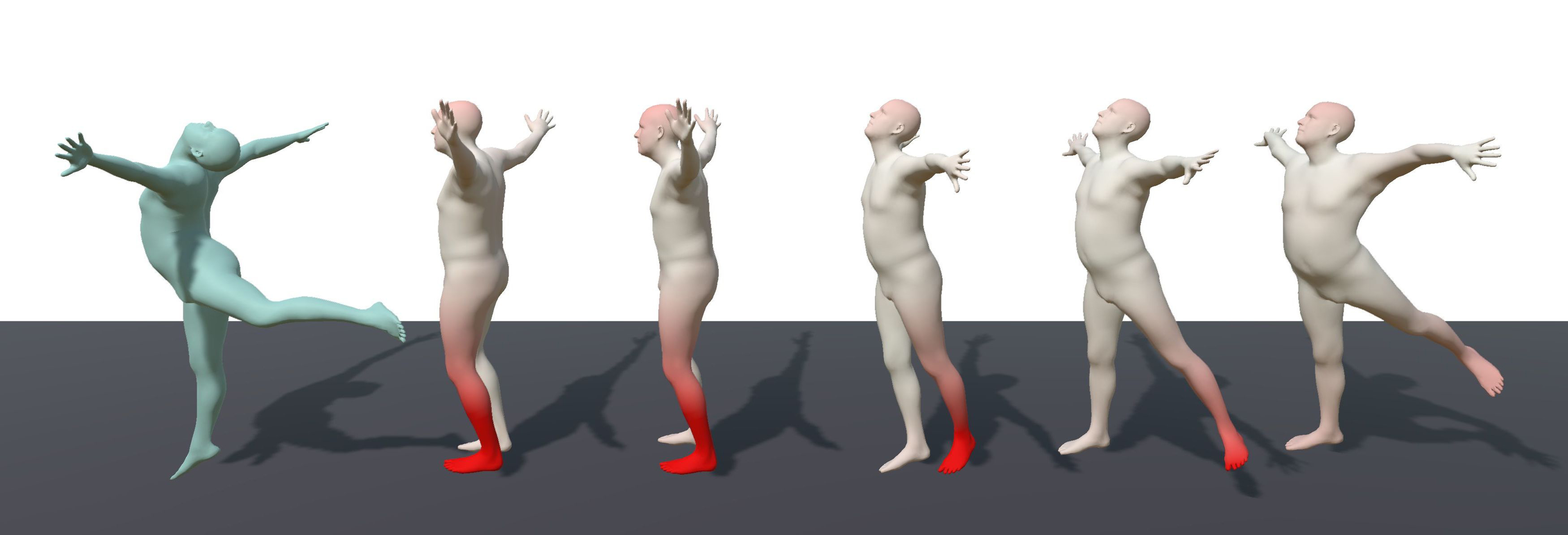}}
\hspace{0.001in}
\subfloat[Walking Motion\label{fig:Competitor:walking}]{%
  \includegraphics[height=0.165\textwidth]{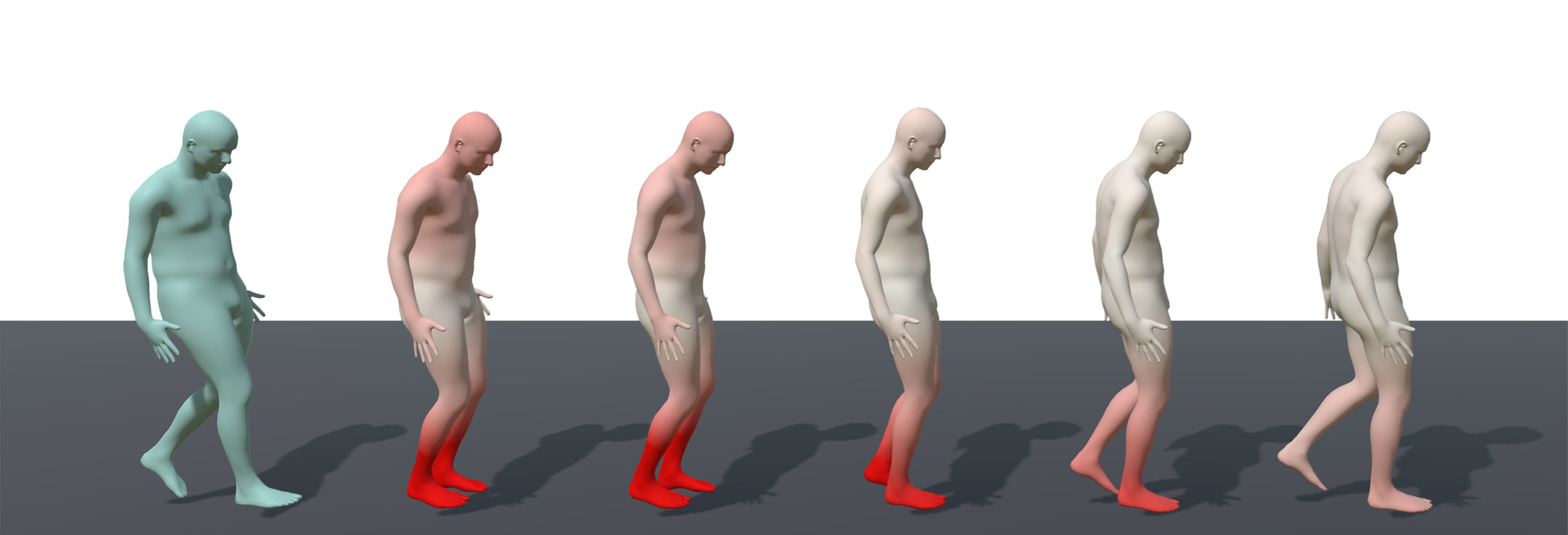}}
\\ \vspace{-0.03in}
\subfloat[Bending Motion\label{fig:Competitor:bending}]{%
  \includegraphics[height=0.165\textwidth]{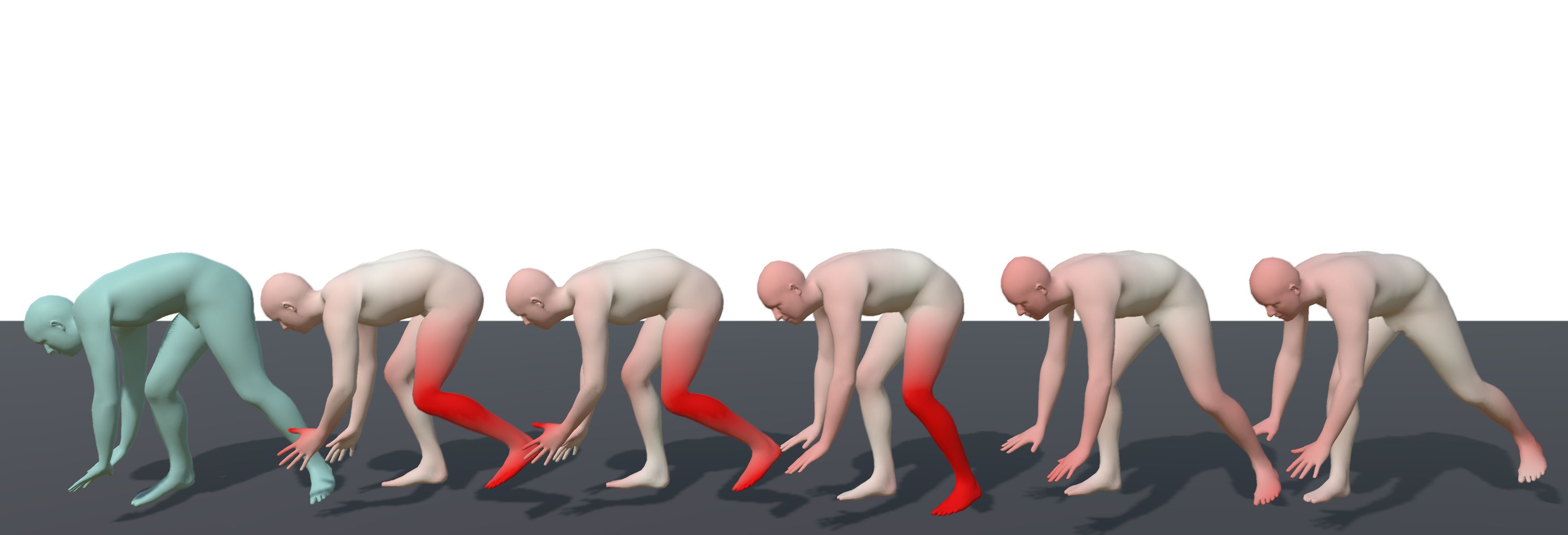}}
\hspace{0.001in}
\subfloat[Kicking Motion\label{fig:Competitor:kicking}]{%
  \includegraphics[height=0.165\textwidth]{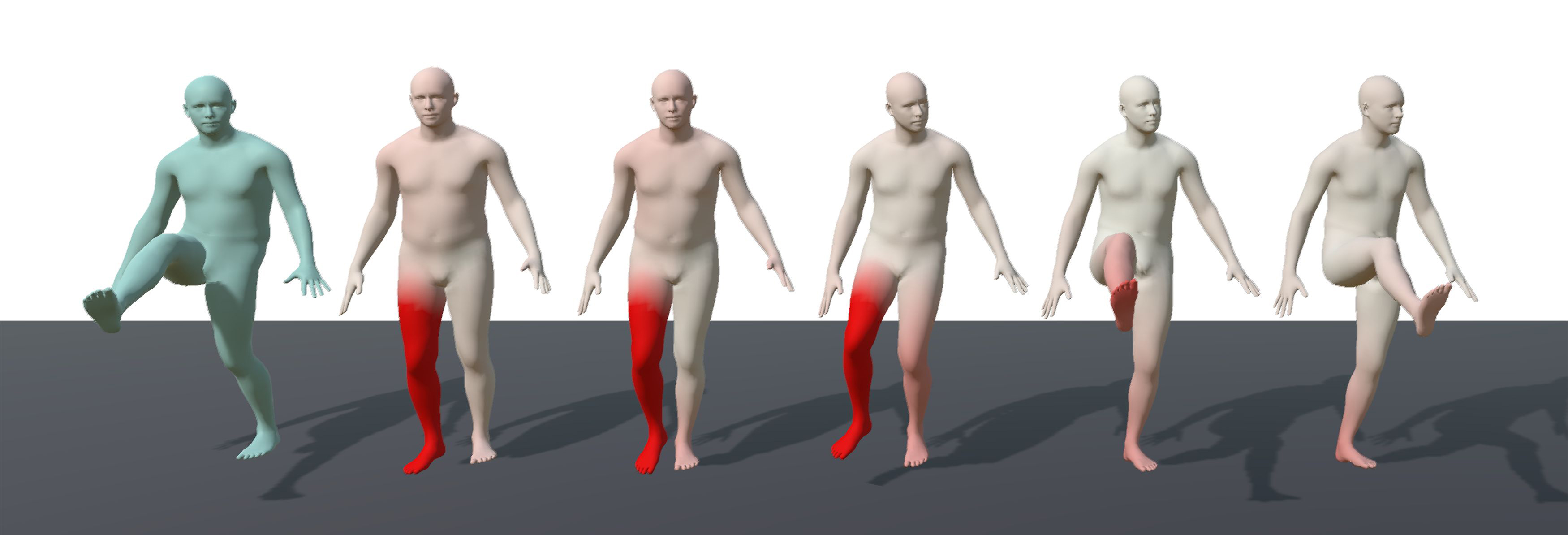}}
\vspace{-0.05in}
\caption{Qualitative results of module-level comparison with state-of-the-art methods. 
The leftmost blue avatar represents the ground truth, while the remaining five avatars from left to right correspond respectively to the methods listed in Rows A–E of Table \ref{tab:Quantitative_DifferentMethods}. Pose errors are color-coded relative to the ground truth, with deeper red indicating greater error.}
\label{fig:Competitor}
\vspace{-0.05in}
\end{figure*}

\vspace{0.15in}
\section{Evaluations}
\label{sec:evaluations}

\vspace{0.05in}
\subsection{Experiments}
\label{subsec:experiments}

We implemented our system and trained all models using an NVIDIA GeForce RTX $3090$ GPU. The training was conducted on our proposed EgoPoseVR dataset synthesized from human motions in the AMASS dataset \cite{mahmood2019amass}. Each dataset was split into 90$\%$ for training and 10$\%$ for testing. The network was optimized using AdamW with the learning rate of $5.0e-4$. The batch size was set to $32$. The overall training objective comprises $2$D joint estimation loss and $3$D pose estimation loss that enforce spatial accuracy and kinematic plausibility. For a complete formulation and loss weighting strategy, please refer to the supplementary material.

In the absence of prior approaches that jointly integrate HMD motion and egocentric visual cues, we compared our EgoPoseVR system with the two state-of-the-art single-modality approaches, i.e., the HMD-based method EgoPoser \cite{jiang2024egoposer} and the egocentric vision-based method EgoPoseFormer \cite{yang2024egoposeformer}, as well as their complementary configurations. To comprehensively evaluate performance, we adopt four standard metrics: MPJPE (Mean Per-Joint Position Error, in cm), PA-MPJPE (Procrustes Aligned MPJPE, in cm), MPJRE (Mean Per-Joint Rotation Error, in degrees), and FPS (Frames Per Second).

\subsection{Comparison with Existing Methods}


To ensure a fair and rigorous comparison with existing approaches, we first adopt EgoPoser \cite{jiang2024egoposer} to reveal the limitations of using HMD motion alone. To incorporate visual cues, we integrate EgoPoseFormer \cite{yang2024egoposeformer} as a refinement module. Notably, EgoPoseFormer alone is excluded as a standalone baseline because it predicts only joint positions rather than rotations, making it incompatible with SMPL-based motion reconstruction, and it depends on ground-truth pelvis-to-camera transformations during training, which are unavailable in VR inference. To overcome these limitations, we provide EgoPoseFormer with pelvis poses predicted by EgoPoser, enabling a fair and applicable integration.

To rigorously assess the effectiveness of each proposed component, we employ a staged evaluation strategy based on controlled module replacement. Starting from \textit{EgoPoser + EgoPoseFormer} baseline, we first replace the motion stream with our HMD encoder while retaining the original visual module, resulting in \textit{Proposed HMD + EgoPoseFormer}. This configuration isolates the contribution of our motion representation under an otherwise identical visual refinement pipeline. We then substitute the visual stream with our RGB-based encoder, forming \textit{Proposed HMD + Proposed RGB (M)}, where (M) denotes the multi-frame input. This intermediate variant highlights the effectiveness of our visual module and ensures a fair comparison with the \textit{EgoPoser + EgoPoseFormer} baseline. Finally, we evaluate our full system (i.e., \textit{EgoPoseVR (HMD + RGB-D (M) + KPO)}), which integrates all proposed modules and achieves superior accuracy in VR-based full-body pose estimation.

\subsubsection{Quantitative Evaluation}

To objectively assess the effectiveness of different methods, we present quantitative comparisons in Table~\ref{tab:Quantitative_DifferentMethods}. Metrics are reported separately for the upper body (-U) and lower body (-L) to better isolate the impact of each module. Additional per-joint metrics are provided in the supplementary material.


As shown in Row B, incorporating visual cues into EgoPoser significantly improves the accuracy of lower-body joint estimation, which primarily depends on visual inference in egocentric settings due to the absence of direct sensing. In Row C, replacing EgoPoser with our proposed HMD module leads to further performance gains across all metrics, with a particularly notable reduction in MPJRE-U. This improvement stems from our dedicated joint embedding design, which more effectively encodes sensor input. Such enhancement is especially valuable for SMPL-based avatar control, where accurate joint rotation is essential for realistic motion synthesis.

To ensure a fair comparison with the baseline in Row B, Row D retains the same input modalities but substitutes EgoPoseFormer with our proposed visual module. Although the accuracy improvement over Row C is relatively modest, the system achieves a substantial increase in FPS, demonstrating a more favorable trade-off between precision and runtime speed. Finally, Row E presents the complete configuration of EgoPoseVR. By incorporating depth input and the KPO module, the system effectively mitigates challenges such as motion blur and occlusion while enforcing kinematic consistency in the upper body. Despite the added computational cost, this full setup delivers the highest accuracy, particularly in upper-body estimation (e.g., MPJPE-U), while maintaining real-time capability, making it well-suited for interactive VR scenarios.

\subsubsection{Qualitative Evaluation}
\label{subsubsec: QualitativeTesting}

The methods corresponding to Rows A–E in Table \ref{tab:Quantitative_DifferentMethods} are visualized under scenarios where the lower body exhibits noticeable motion, as shown in Fig. \ref{fig:Competitor}. EgoPoser infers lower-body motion primarily from contralateral arm–leg coordination. However, when arm swings are not evident (e.g., Fig. \ref{fig:Competitor:walking}), its predictions become unreliable. The approaches in Rows B–C, which build on EgoPoseFormer, depend heavily on the pelvis position within the image view. Consequently, when the user makes large head movements upward (Fig. \ref{fig:Competitor:leaping}), downward (Figs. \ref{fig:Competitor:walking} and \ref{fig:Competitor:bending}) or to the side (Fig. \ref{fig:Competitor:kicking}), the predicted lower-body poses are unreliable.

In contrast, the proposed RGB module (Row D) leverages visible joints in the image to enhance the HMD-based estimation of lower-body joints, without depending on pelvis localization, thereby accommodating larger viewpoint variations. Incorporating depth information and KPO further improves performance by enhancing $3$D spatial understanding of the lower body and enforcing upper-body constraints based on HMD sensor input, resulting in more accurate and robust full-body pose estimation.

\subsection{Ablation Study}

We conduct an ablation study on both input modalities and model components to analyze their individual contributions (Table~\ref{tab:ablationstudy}).

For input modalities, starting from an HMD-only baseline, we first add single-frame RGB input, which contributes per-frame visual features without temporal context. This yields a marked improvement in positional accuracy, particularly for lower-body joints where HMD signals alone provide weak kinematic constraints. We then replace single-frame input with a temporally stacked RGB sequence, allowing the network to exploit short-term motion continuity. This further reduces both positional noise and orientation instability, especially under abrupt head movement. The inclusion of depth in the multi-frame configuration brings additional gains across all metrics by providing geometric cues that disambiguate joint positions under occlusion and motion blur.

\begin{table}[h]
\small
\centering
\caption{Ablation study of input modalities and model components. (S): single-frame RGB input; (M): multi-frame RGB input.
Row colors indicate accuracy improvement from the previous row:
\textcolor{lightyellow1}{\textbf{Yellow}} (\textless10\%),
\textcolor{darkgreen}{\textbf{Light Green}} (10–20\%),
\textcolor{darkgreen1}{\textbf{Dark Green}} (\textgreater20\%).}
\vspace{-0.05in}
\renewcommand{\arraystretch}{1.0}
\setlength\tabcolsep{1pt}
\begin{tabular}{p{28mm}| C{13mm}  C{16mm} C{13mm} C{11mm}}
\hlineB{3}
\centering Method & MPJPE $\downarrow$ & PA-MPJPE $\downarrow$ & MPJRE $\downarrow$ & FPS $\uparrow$\\
\hline
\hline
HMD Only & $6.26 \pm 4.03$ & $4.85 \pm 3.05$ & $9.26 \pm 3.85$ & $212 \pm 6$\\

HMD + RGB (S) & \cellcolor{darkgreen}$4.42 \pm 3.47$ & \cellcolor{darkgreen}$3.37 \pm 2.57$ & \cellcolor{lightgreen}$8.12 \pm 3.82$ & $158 \pm 8$\\

HMD + RGB (M) & \cellcolor{lightyellow}$4.25 \pm 3.28$ & \cellcolor{lightyellow}$3.22 \pm 2.47$ & \cellcolor{lightyellow}$7.90 \pm 3.68$ & $155 \pm 6$\\

HMD + RGB-D (M) & \cellcolor{lightyellow}$3.95 \pm 3.05$ & \cellcolor{lightyellow}$2.96 \pm 2.21$ & \cellcolor{lightyellow}\textbf{$7.61 \pm 3.51$} & $154 \pm 7$\\
\hline
\; + KPO (Full System) & \cellcolor{darkgreen}\textbf{2.92 $\pm$ 2.16} & \cellcolor{lightyellow}\textbf{2.70 $\pm$ 2.17} & \textbf{7.61 $\pm$ 3.51} & \textbf{97 $\pm$ 6}\\

\; -- Cross-Modal Fusion & $4.08 \pm 3.03$ & $3.10 \pm 2.19$ & $7.65 \pm 3.62$ & $156 \pm 5$\\
\; -- Spatiotemporal Encoder & $6.42 \pm 4.19$ & $5.35 \pm 3.28$ & $14.29 \pm 5.07$ & $189 \pm 6$ \\

\hlineB{3}
\end{tabular}
\label{tab:ablationstudy}
\vspace{-0.05in}
\end{table}

For model components, the proposed KPO module in Sec. \ref{subsec:skeletonoptimization} is further applied on top of \textit{HMD + RGB-D (M)}, and particularly improves full-body joint position consistency while preserving predicted joint orientations. Although the refinement introduces computational cost, the final system still achieves $97$ FPS, ensuring its practicality for real-time VR interactions. Additional ablation study results are provided in the supplementary material. To assess the necessity of the spatiotemporal encoder and the cross-modal fusion module, we remove each component from the full system. While this expectedly improves inference speed, it leads to reduced full-body joint pose accuracy. Replacing cross-modal fusion with simple feature concatenation results in only a modest performance drop, as modality-specific features are already effectively encoded by the spatiotemporal encoder before fusion.



\begin{figure}[h]
\vspace{-0.05in}
  \centering
  \setlength{\tabcolsep}{1pt}
  \renewcommand{\arraystretch}{1.0}

  \begin{tabular}{@{}c ccc@{}} 
    \raisebox{0.03in}{\rotatebox{90}{\scriptsize{Ground Truth}}} &
    \includegraphics[width=0.12\textwidth]{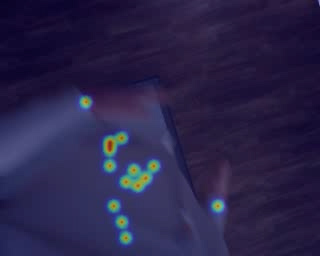} & \hspace{0.0002in}
    \includegraphics[width=0.12\textwidth]{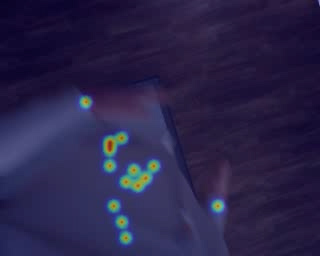} & \hspace{0.0002in}
    \includegraphics[width=0.12\textwidth]{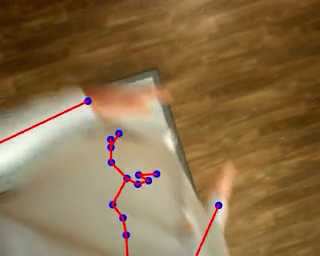} \\
    \raisebox{0.10in}{\rotatebox{90}{\scriptsize{RGB Input}}} &
    \includegraphics[width=0.12\textwidth]{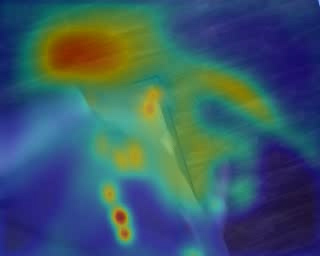} & \hspace{0.0002in}
    \includegraphics[width=0.12\textwidth]{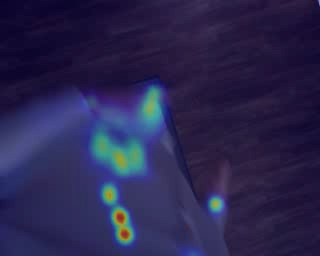} & \hspace{0.0002in}
    \includegraphics[width=0.12\textwidth]{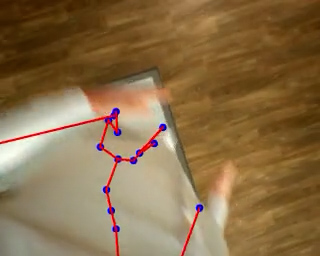} \\
    \raisebox{0.03in}{\rotatebox{90}{\scriptsize{RGB-D Input}}} &
    \includegraphics[width=0.12\textwidth]{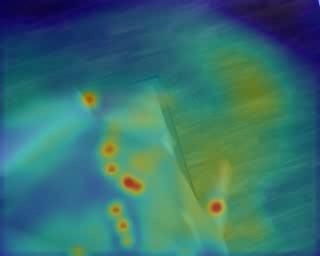} & \hspace{0.0002in}
    \includegraphics[width=0.12\textwidth]{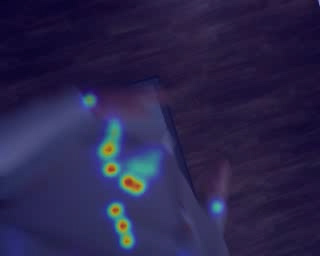} & \hspace{0.0002in}
    \includegraphics[width=0.12\textwidth]{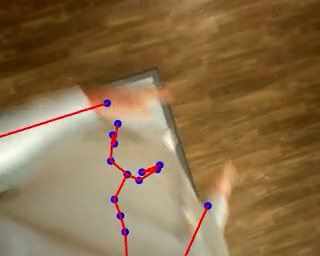} \\
    & \scriptsize Without $\boldsymbol{\zeta}$ &
      \scriptsize With $\boldsymbol{\zeta}$ &
      \scriptsize Pose Re-Projection \\
  \end{tabular}
  \vspace{-0.05in}
  \caption{Effect of visual input modality and the proposed visibility probability factor $\zeta$ on heatmap quality and the resulting 3D pose estimation, evaluated via 2D re-projections of the predicted joints.}
    \vspace{-0.05in}
  \label{fig:heatmap_reproj_grid}
\end{figure}

\begin{figure*}[t]
\centering
\begin{minipage}[c]{0.03\textwidth}
  \centering
  \vspace{-0.25cm}
  \rotatebox{90}{Virtual Avatars} \\[1.22cm]
  \rotatebox{90}{Real-world Users}     
\end{minipage}
\begin{minipage}{0.31\textwidth}
  \centering
  \includegraphics[width=\linewidth]{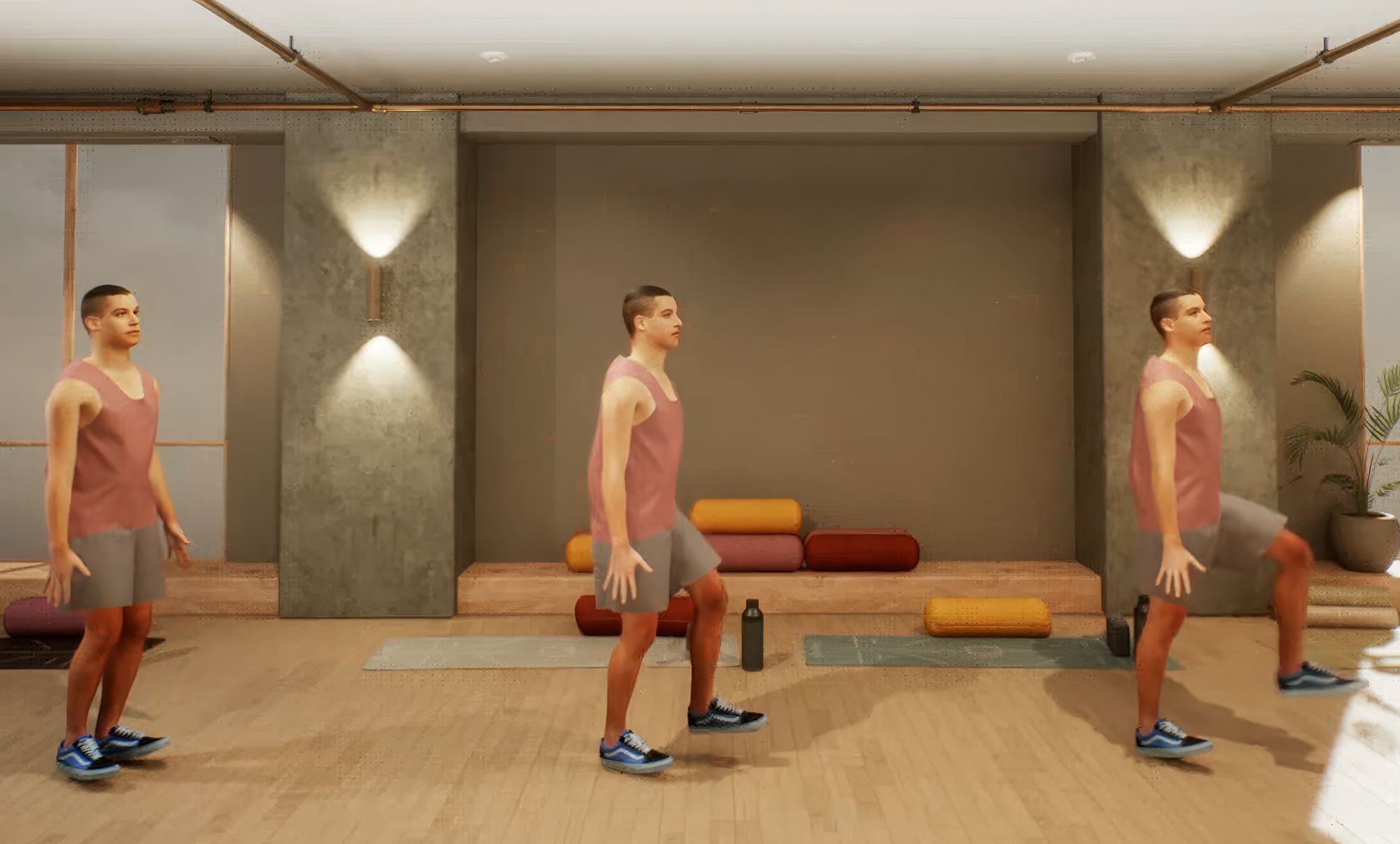}\\ \vspace{0.03in}
  \includegraphics[width=\linewidth]{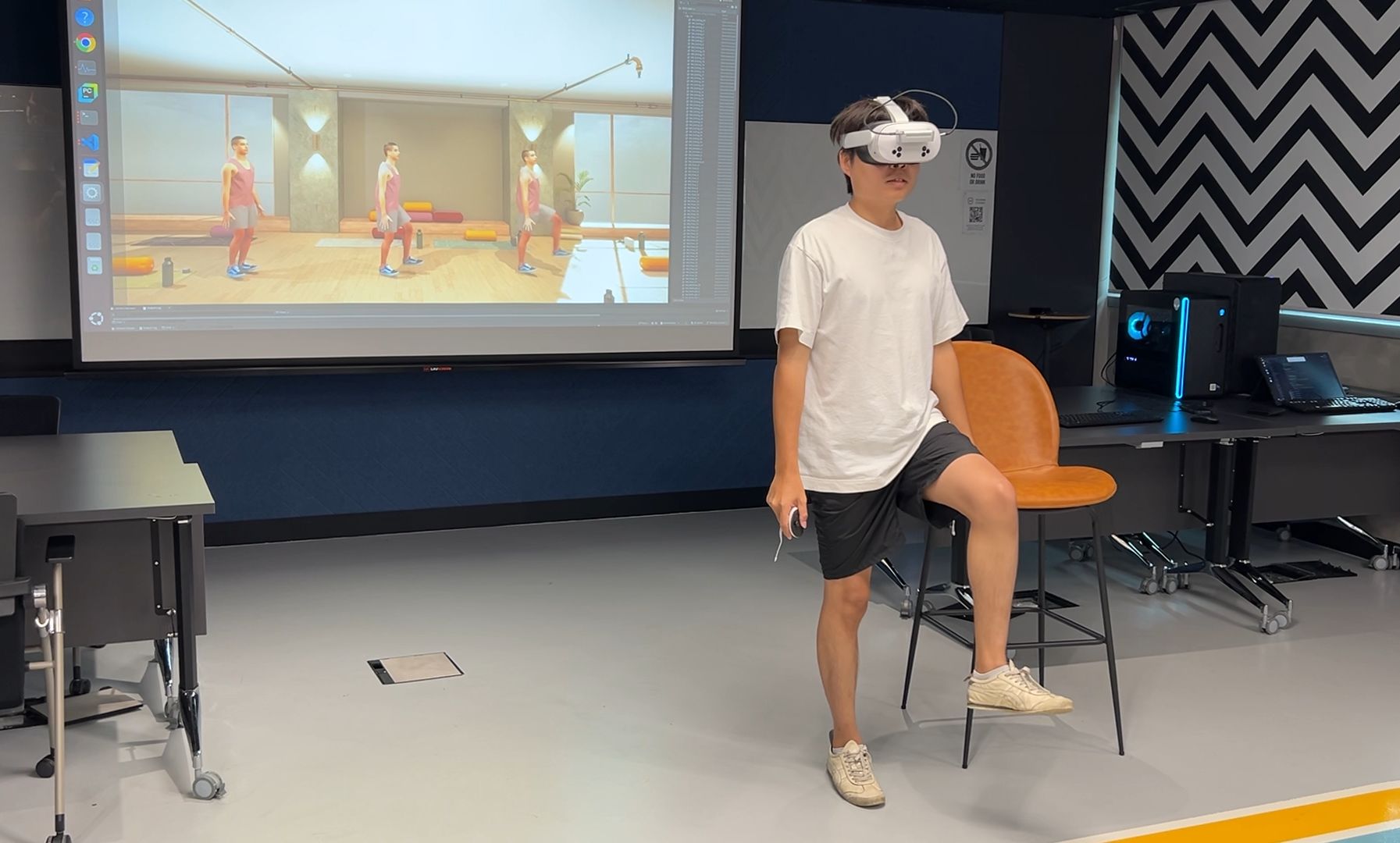}\\
  \subcaption{High-Knee Motion}
  \label{fig:Competitor:highknee}
\end{minipage}\hspace{0.03in}
\begin{minipage}{0.31\textwidth}
  \centering
  \includegraphics[width=\linewidth]{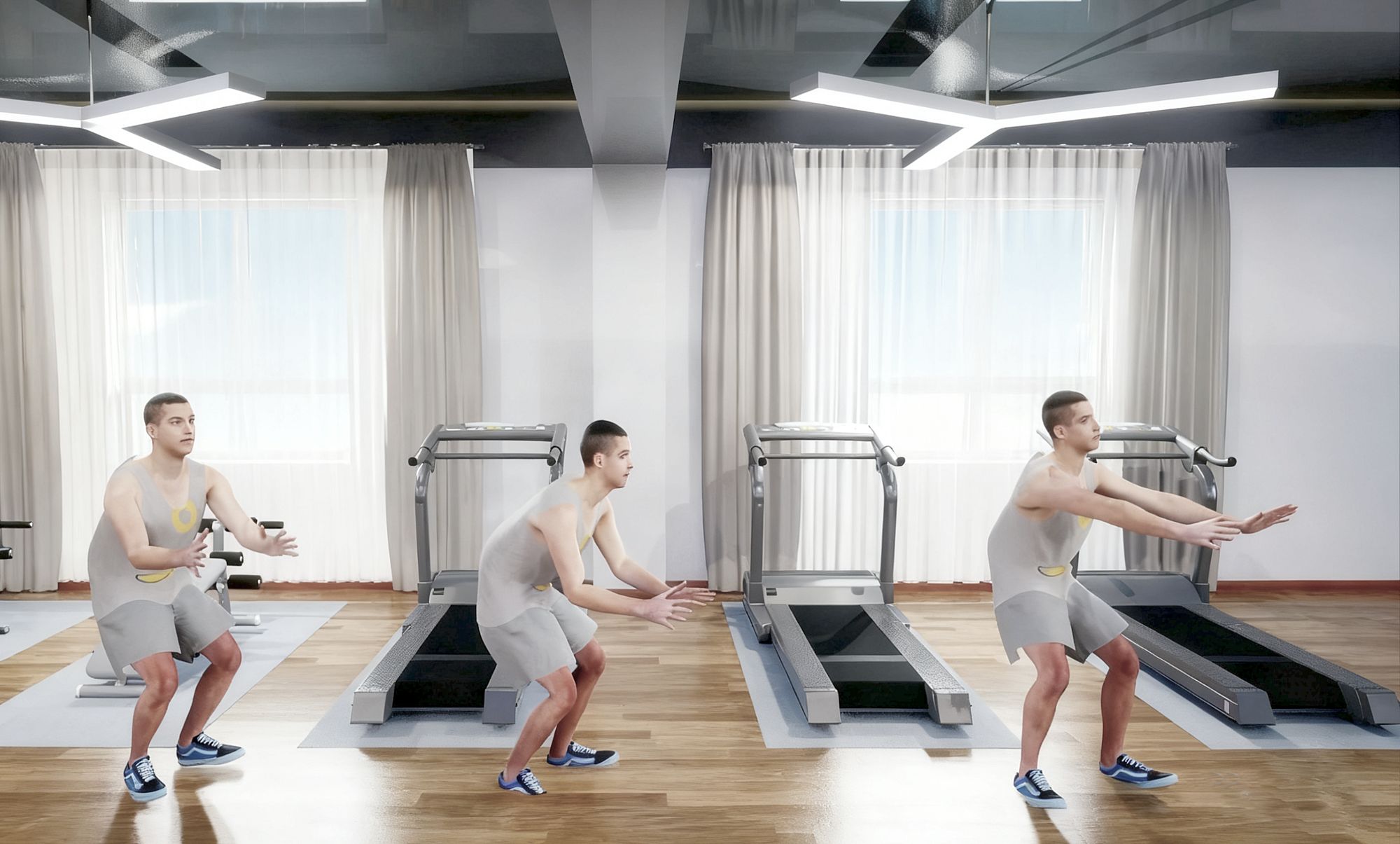}\\ \vspace{0.03in}
  \includegraphics[width=\linewidth]{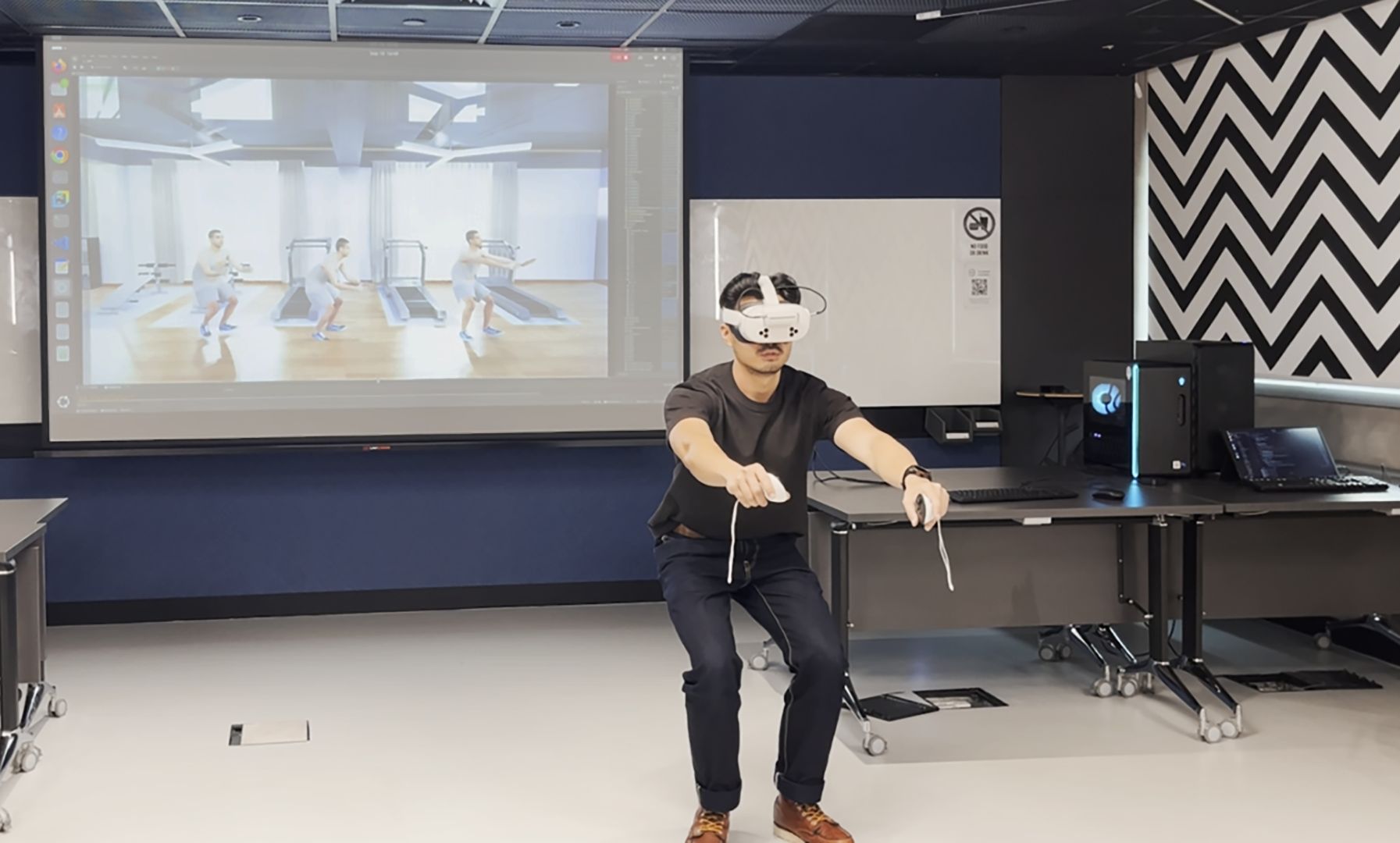}\\
  \subcaption{Squat Motion}
  \label{fig:Competitor:squat}
\end{minipage}\hspace{0.03in}
\begin{minipage}{0.31\textwidth}
  \centering
  \includegraphics[width=\linewidth]{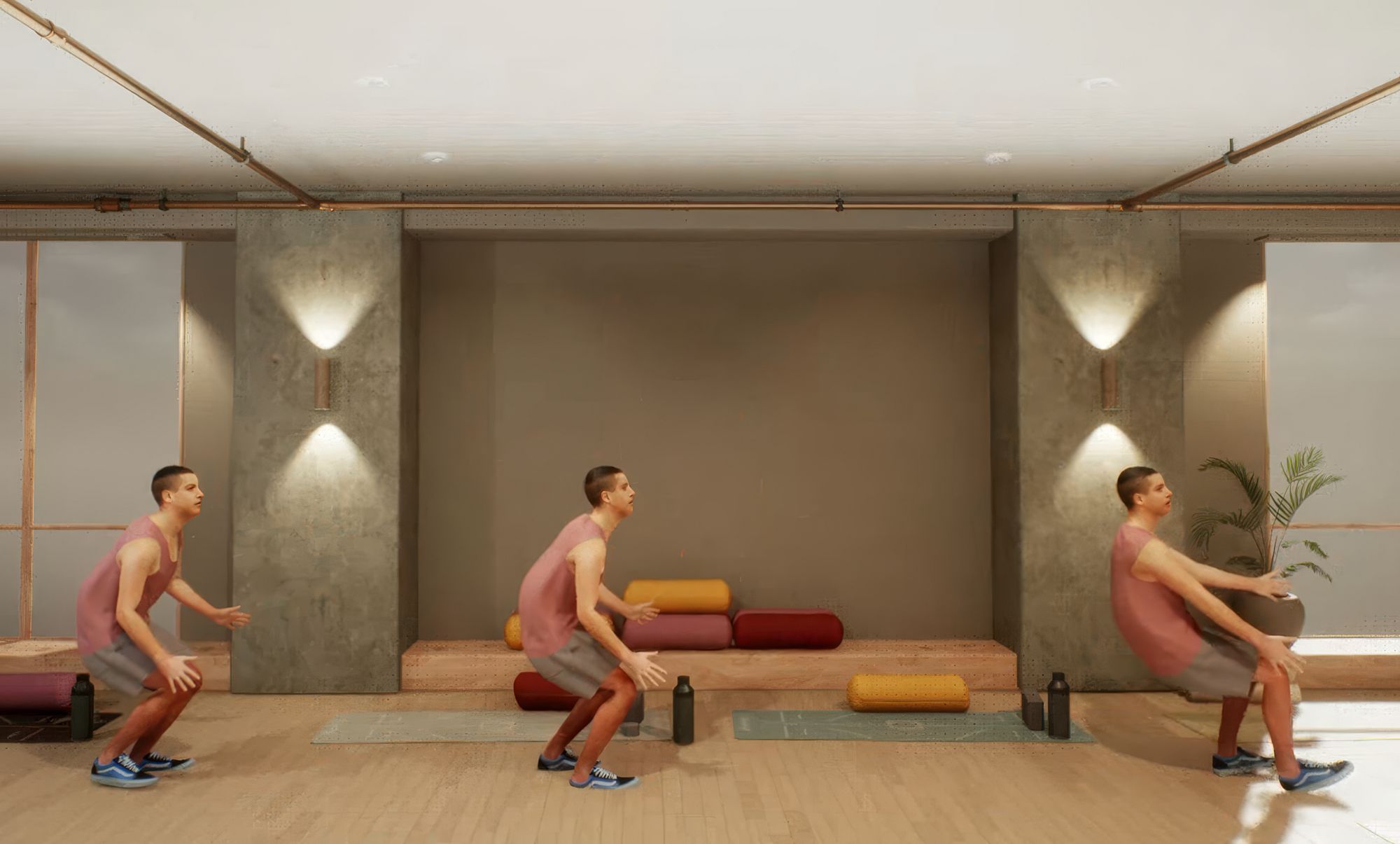}\\ \vspace{0.03in}
  \includegraphics[width=\linewidth]{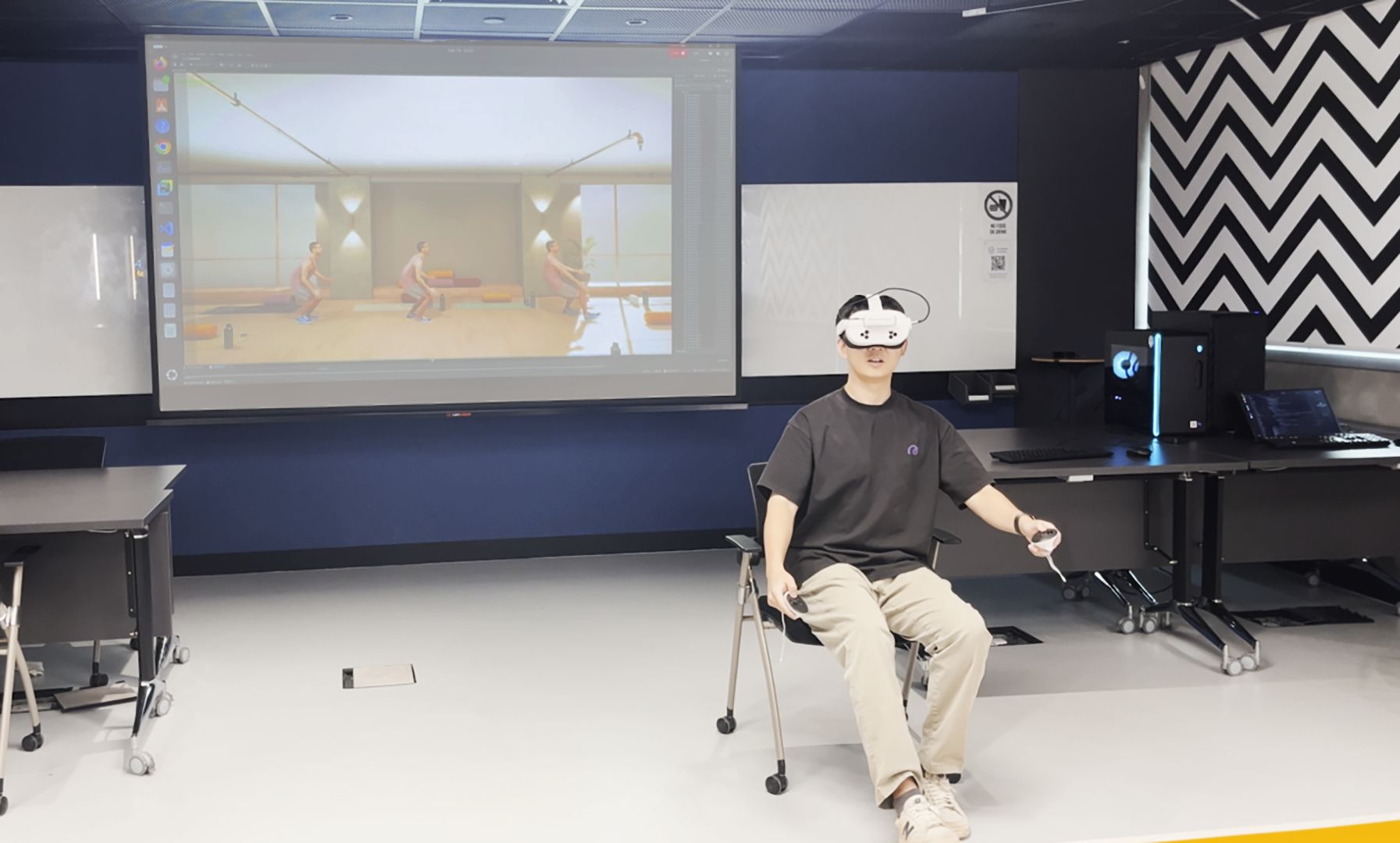}\\
  \subcaption{Sitting Motion}
  \label{fig:Competitor:sitting}
  \end{minipage}
  \vspace{-0.10in}
\caption{Comparison with existing methods in the real-world scenes. From left to right, the avatars are generated by EgoPoser \cite{jiang2024egoposer}, EgoPoser \cite{jiang2024egoposer} + EgoPoseFormer \cite{yang2024egoposeformer}, and the proposed EgoPoseVR, respectively. The foreground users illustrate the corresponding real-world motions, while the background avatars demonstrate their synchronized virtual counterparts in real time.}
\label{fig:usdemo}
\end{figure*}

\vspace{-0.05in}
\subsection{Heatmap-Driven Joint Accuracy Analysis}

To examine how different factors affect $2$D heatmap quality, we compare input configurations across two dimensions: image modality and the integration of the proposed visibility probability factor $\zeta$ in Sec. \ref{subsubsec:visualstream}. For quantitative validation, the predicted $3$D joints are re-projected onto the original $2$D image plane via known camera intrinsics, allowing a direct assessment of spatial alignment with the input observations.



The leftmost column of Fig.~\ref{fig:heatmap_reproj_grid} shows that without incorporating $\zeta$, the heatmaps corresponding to out-of-view joints introduce substantial noise and artifacts. This occurs because fixed-threshold filtering cannot adaptively suppress heatmap responses from invisible joints. To address this limitation, we introduce a joint visibility head that predicts the probability of each joint being within the visible region. As shown in the middle column, integrating $\zeta$ effectively suppresses noisy heatmap signals and sharpens the heatmap distributions, thereby focusing on the valid joint regions. By comparing the RGB and RGB-D rows in Fig.~\ref{fig:heatmap_reproj_grid}, it is evident that depth cues enable the network to better capture the 3D spatial structure of the scene, resulting in more accurate joint localization. This refinement ensures temporally stable 2D joint sequences, which serve as reliable input to the spatiotemporal encoder and ultimately lead to reliable 3D pose estimation.


\subsection{User Study in Real-life Environments}

To evaluate the effectiveness of different approaches in the real-world scenes, we conducted a user study comparing EgoPoser \cite{jiang2024egoposer} (EP), EgoPoser \cite{jiang2024egoposer} + EgoPoseFormer \cite{yang2024egoposeformer} (EP+EF), and our proposed EgoPoseVR, i.e., Row A, B and E in Table \ref{tab:Quantitative_DifferentMethods}. The study protocol was reviewed and approved by the Institutional Review Board (IRB) of our university. All compared models were trained using the proposed dataset. Building on the cross-device data transmission design described in Sec. \ref{sec:DataTransmission}, to ensure a fair and consistent comparison, virtual avatars generated by the different methods are simultaneously rendered side by side in UE render and driven by identical user input from the same VR device.

\begin{table}[h]
\footnotesize
\centering
\caption{Post-session questionnaire used for the subjective evaluation of different methods.}
\renewcommand{\arraystretch}{1.0}
\setlength\tabcolsep{4pt}
\begin{tabular}{M{11mm}| L{65mm}}
\hlineB{3}
\textbf{Question} & \multicolumn{1}{c}{\textbf{Description}}  \\
\hline
\hline
Q1 & The avatar’s upper body movements accurately reflected my real-world movements. \\ \hline
Q2 & The avatar’s lower body movements accurately reflected my real-world movements. \\ \hline
Q3 & The avatar’s movements were synchronized (real-time) with my actions in real time. \\ \hline
Q4 & The avatar’s motion appeared stable and free of noticeable jittering. \\ \hline
Q5& I felt that the avatar’s body was a natural extension of my own body. \\ \hline
Q6& I would prefer to use this pose estimation method and its associated device in future VR applications. \\
\hlineB{3}
\end{tabular}
\vspace{-0.05in}
\label{tab:userstudy_questionnaire}
\end{table}

A total of $20$ participants were recruited from the university staff and students (mean age = $24.9$ years, standard deviation = $3.73$), consisting of $11$ males and $9$ females. After reading and signing the consent form, participants first completed a demographic questionnaire to record their basic information. Participants were informed of the general purpose during the consent process but were blind to the specific study hypotheses. Participants exhibited heterogeneous prior VR experience, with $14$ participants reporting prior VR exposure and $6$ reporting none. In addition, $8$ participants also reported prior experience with pose tracking or related development tasks. Following a short demonstration of our system, they were asked to wear the headset and freely perform basic whole-body movements (e.g., walking, sitting, leg raising, and bending) within $5$ mins. After the session, participants filled out the 5-point Likert-scale questionnaire shown in Table~\ref{tab:userstudy_questionnaire}. Based on standardized embodiment questionnaires \cite{gonzalez2018avatar,borrego2019embodiment}, our questionnaire was designed to assess four dimensions of the system: \textit{tracking accuracy} (Q1–Q2), \textit{responsiveness and stability} (Q3–Q4), \textit{sense of embodiment} (Q5), and \textit{intention for future use} (Q6). We adapted items most relevant to the system’s functional evaluation rather than adopting a full embodiment inventory. During the experiment, participants were not informed of the arrangement of avatars corresponding to different methods in the UE environment, ensuring unbiased evaluation.

\begin{figure}[h]
\centering
\hspace{-0.10in}
\includegraphics[width=0.49\textwidth]{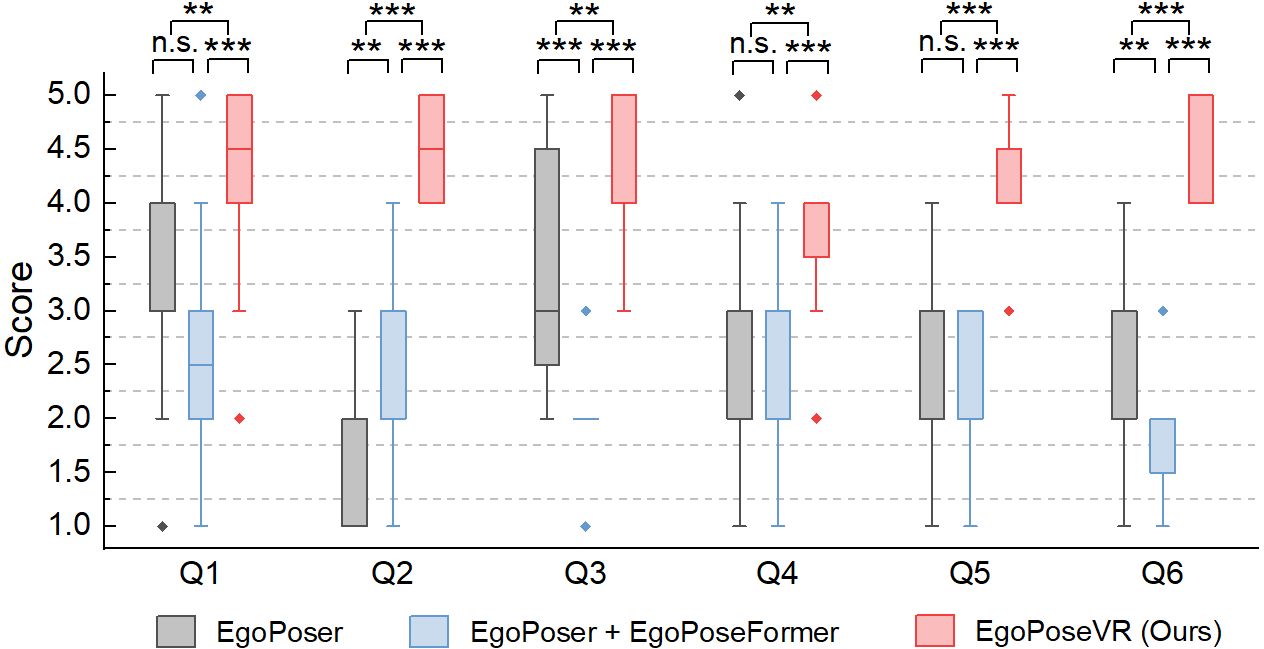}
\caption{Statistical analysis of the user study. The box represents the interquartile range (IQR), the central line indicates the median. Outliers are shown as individual points. (n.s.): no statistical significance, (*): p $\leq$ 0.05, (**): p $\leq$ 0.01, (***): p $\leq$ 0.001.}
\label{fig:uschartbox}
\vspace{-0.05in}
\end{figure}

The questionnaire ratings were collected on a 5-point Likert scale, yielding ordinal data that cannot be assumed to follow a normal distribution. Accordingly, we employed non-parametric analyses, applying Friedman tests to assess overall differences across methods and Wilcoxon signed-rank tests with the Holm–Bonferroni adjustment for pairwise comparisons. The Friedman test results revealed significant differences for all questionnaire items ($p < 0.001$). 
Post hoc tests further showed that EgoPoseVR yielded significantly higher rating than both EP and EP+EF ($p < 0.01$), as shown in Fig. \ref{fig:uschartbox}. These findings demonstrate that participants consistently perceived evident differences among the three models across all evaluation aspects. Detailed statistical results are provided in the supplementary material.


The Friedman test indicated a significant effect for Q1 (\textit{upper-body tracking accuracy}) ($\chi^2(2) = 21.794$, $p < 0.0001$, Kendall’s W = 0.545). When comparing EP+EF and EP, EP+EF employed the same HMD-based upper-body estimation strategy as EP, with no significant difference ($p=0.051$). However, the additional latency introduced by EP+EF reduced perceived accuracy relative to EP. For \textit{lower-body tracking accuracy} (Q2), the Friedman test also showed a significant effect for Q2 ($\chi^2(2) = 35.368$, $p < 0.0001$, Kendall’s W = 0.884). EP performed worst as it relied solely on HMD motion, while EP+EF offered moderate improvements but still struggled with large joint-angle poses such as single-leg squats and knee lifts (Fig. \ref{fig:usdemo}). In contrast, EgoPoseVR enhanced by the dedicated HMD and KPO modules, yielded superior upper-body accuracy and significantly outperformed both baselines in lower-body motions (all $p < 0.001$), owing to the spatiotemporal encoder and cross-modal module. 

For \textit{responsiveness and stability} (Q3–Q4), significant condition effects were observed 
(Q3: $\chi^2(2) = 31.886$, $p < 0.001$, Kendall’s $W = 0.797$; 
Q4: $\chi^2(2) = 25.323$, $p < 0.001$, Kendall’s $W = 0.633$). EP+EF’s computational overhead hindered timely avatar responses, resulting in markedly lower responsiveness scores. In terms of stability, EP exhibited jitter from its heuristic propagation of lower-body poses from upper-body motion, while EP+EF displayed frame-to-frame inconsistency amplified by latency, with no statistically significant difference between the two conditions ($p = 0.749$). The proposed method alleviated both issues (all $p < 0.001$), though when the arms occluded the lower body or when parts moved outside the field of view, reconstructed motion was prone to becoming less smooth. 

For \textit{sense of embodiment} (Q5) and \textit{intention for future use} (Q6), significant condition effects were observed 
(Q5: $\chi^2(2) = 31.521$, $p < 0.001$, Kendall’s $W = 0.788$; 
Q6: $\chi^2(2) = 34.274$, $p < 0.001$, Kendall’s $W = 0.857$). Post hoc tests showed that participants rated the proposed method significantly higher than EP and EP+EF (all $p < 0.001$). In term of Q5, the lack of a significant difference between EP and EP+EF ($p = 0.090$) indicates that partial or unstable lower-body tracking is insufficient to substantially improve embodiment. Stronger embodiment arose from superior tracking accuracy and temporal stability, which also translated into a greater willingness to adopt the system in future VR applications, showing that subjective acceptance was consistent with improvements in the preceding technical questions.

\section{Limitations and Future Work}
\label{sec:limitations}

\begin{figure}[h]
\centering
\includegraphics[width=0.23\textwidth]{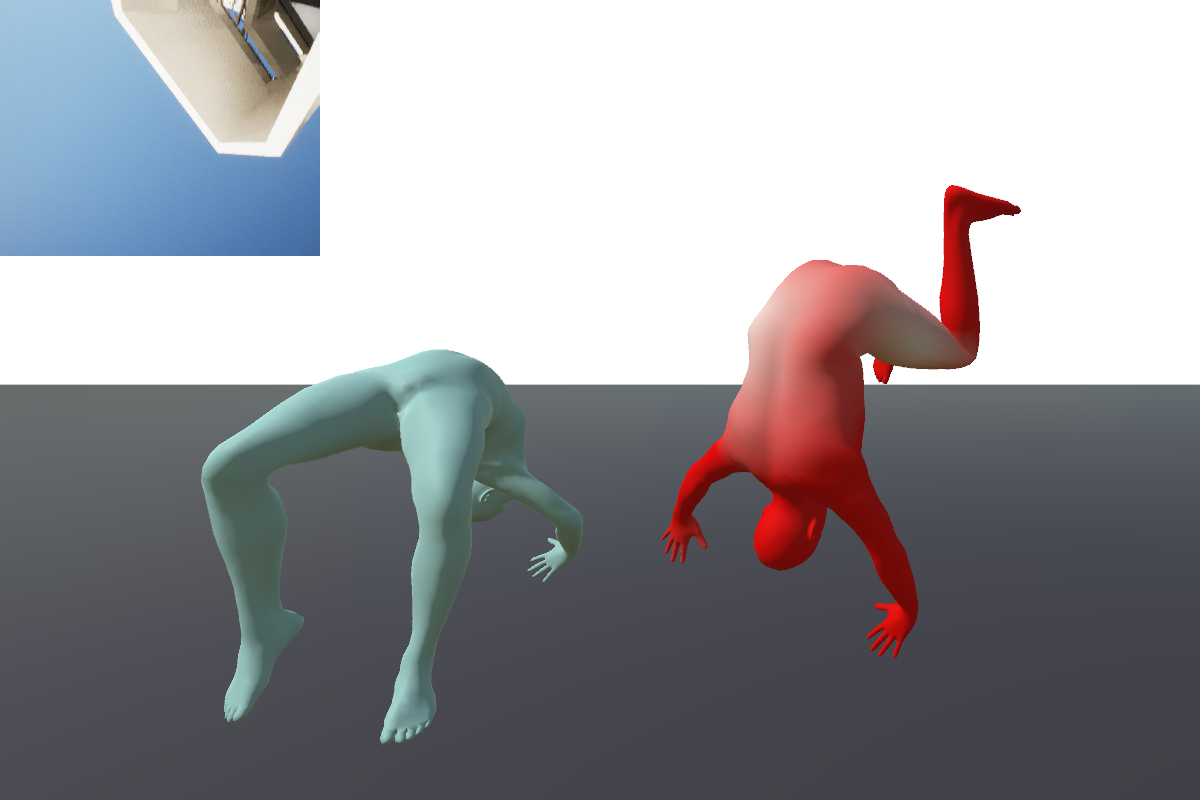} 
\includegraphics[width=0.23\textwidth]{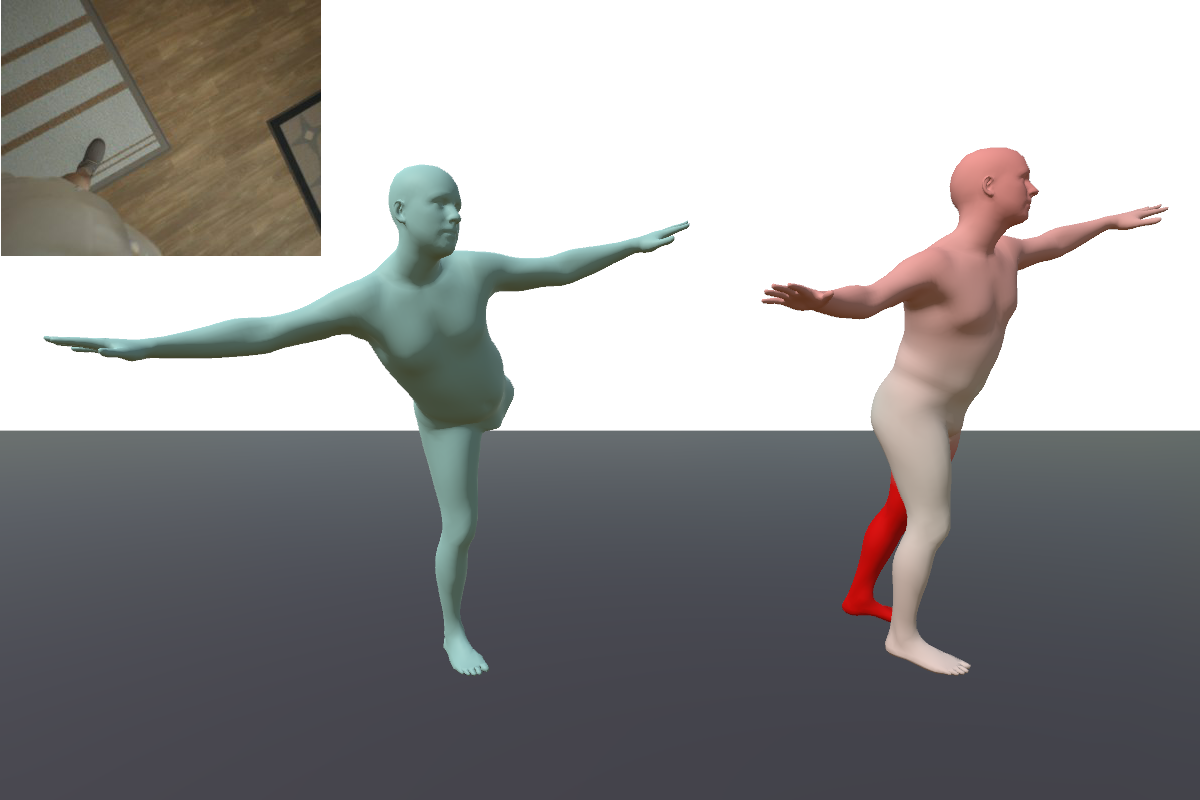}
\vspace{-0.05in}
\caption{Failure cases under partial or complete observability loss in egocentric views. The blue avatar denotes the ground truth, and the right avatar shows the predicted pose with color-coded errors.}
\label{fig:limitation}
\vspace{-0.05in}
\end{figure}

Our method has several limitations that open avenues for future exploration. First, the effectiveness of the RGB-D refinement module relies on the visibility of the user body within the camera's view. In cases where the user body is outside the field of view or severely occluded (Fig. \ref{fig:limitation}), the benefit of cross-modal fusion diminishes. Future work may consider incorporating temporal priors or generative models to better handle such visually sparse conditions. Second, although the avatar motions in our dataset are driven by real-life motion capture data, the corresponding visual observations are synthesized, which inevitably introduces a sim-to-real gap. To mitigate this issue, we have incorporated visibility-aware modeling (Sec. \ref{subsubsec:visualstream}) and kinematic alignment with HMD sensor measurements (Sec. \ref{subsec:skeletonoptimization}) to reduce sensitivity to complex motions and environment scenes. Nevertheless, future work will focus on collecting real-world, temporally synchronized HMD motion and RGB-D data across diverse physical environments to further improve robustness under realistic deployment conditions. Third, while SMPL enables efficient prediction and animation, it does not incorporate biomechanical constraints. As a result, the generated poses may lack physical plausibility or high-frequency muscular dynamics. Extending our framework to incorporate musculoskeletal or physics-based priors could lead to more anatomically realistic pose reconstruction.

\vspace{-0.10in}
\section{Conclusion}
\label{sec:conclusions}

We present \textit{EgoPoseVR}, a unified framework for egocentric full-body pose estimation in VR, leveraging spatially sparse motion trajectories from HMD and spatial observations from a headset-mounted downward-facing RGB-D camera. Our method introduces a spatiotemporal fusion strategy that refines pose estimates derived from HMD motion input using visibility-aware RGB-D features, with a particular focus on improving lower-body accuracy under occlusion and limited viewpoints. To support this task, we construct a large-scale, VR-specific dataset featuring temporally synchronized RGB-D and HMD motion data, enabling effective learning under realistic egocentric conditions. Extensive evaluations across synthetic and real-world VR scenarios demonstrate that \textit{EgoPoseVR} achieves higher accuracy in both joint positions and rotations than state-of-the-art egocentric pose estimation baselines, while maintaining real-time performance. It offers portable and infrastructure-free full-body tracking without the need for additional body-worn sensors. These findings underscore its promise for facilitating natural avatar embodiment and seamless interaction in rehabilitation training and other immersive VR applications.

\bibliographystyle{abbrv-doi}

\bibliography{Reference}
\end{document}